\documentclass{article} 
\usepackage{iclr2027_conference,times}

\usepackage{booktabs}
\usepackage{multicol}
\usepackage{multirow}
\usepackage{amsmath,amsfonts,bm}
\usepackage{xspace}

\usepackage[table,dvipsnames]{xcolor}

\usepackage{microtype}
\usepackage{cuted}
\usepackage{enumitem}
\usepackage{pifont}

\usepackage[most]{tcolorbox}
\usepackage{titletoc}
\usepackage{url}
\usepackage{subcaption}
\usepackage{tcolorbox}
\usepackage{wrapfig}

\usepackage[
    colorlinks=true,
    citecolor=BlueViolet,
    linkcolor=purple,
    urlcolor=BlueViolet
]{hyperref}

\newcommand\DoToC{%
  \startcontents
  \printcontents{}{1}{\noindent \textbf{\Large{Table of Contents in Appendix}}\vskip3pt\vskip5pt}
  \vskip3pt\vskip5pt
}

\newcolumntype{L}{>{$}l<{$}}
\newcolumntype{R}{>{$}r<{$}}
\newcolumntype{C}{>{$}c<{$}}

\renewcommand{\paragraph}[1]{\vspace{.4em}\noindent\textbf{#1.}}

\def\eqref#1{equation~\ref{#1}}

\def\1{\bm{1}}

\def\vb{{\bm{b}}}

\def\vg{{\bm{g}}}
\def\vh{{\bm{h}}}

\def\vu{{\bm{u}}}
\def\vv{{\bm{v}}}
\def\vw{{\bm{w}}}

\def\vz{{\bm{z}}}

\def\mH{{\bm{H}}}

\def\mW{{\bm{W}}}

\DeclareMathAlphabet{\mathsfit}{\encodingdefault}{\sfdefault}{m}{sl}
\SetMathAlphabet{\mathsfit}{bold}{\encodingdefault}{\sfdefault}{bx}{n}

\newcommand{\class}{\texttt{[CLS]}\xspace}
\newcommand{\method}{\textsc{RecSAEI}\xspace}

\usepackage{hyperref}
\usepackage{url}

\title{From Activation to Intervention:\\ Interpreting Sparse Visual Features through Reconstruction}

\author{Jianyang Gu$^{1,2\dagger}$\; Jacob Beattie$^1$\; Mridul Khurana$^3$\; Julia Chae$^2$\; Xi Xiao$^4$\\
\textbf{Cheng Zhang$^5$\; Anuj Karpatne$^3$\; Wei-Lun Chao$^6$\; Sara Beery$^{2\dagger}$\; Tanya Berger-Wolf$^{1\dagger}$}\\[3pt]
{\small $^1$The Ohio State University\quad $^2$Massachusetts Institute of Technology\quad $^3$Virginia Tech}\\
{\small $^4$University of Alabama, Birmingham\quad $^5$Texas A\&M University\quad $^6$Boston University}\\[3pt]
{\small \texttt{\{gu.1220,berger-wolf.1\}@osu.edu\; beery@mit.edu}} 
}

\iclrfinalcopy 
\begin{document}

\maketitle

\begin{figure}[h]
    \centering
    \includegraphics[width=\linewidth]{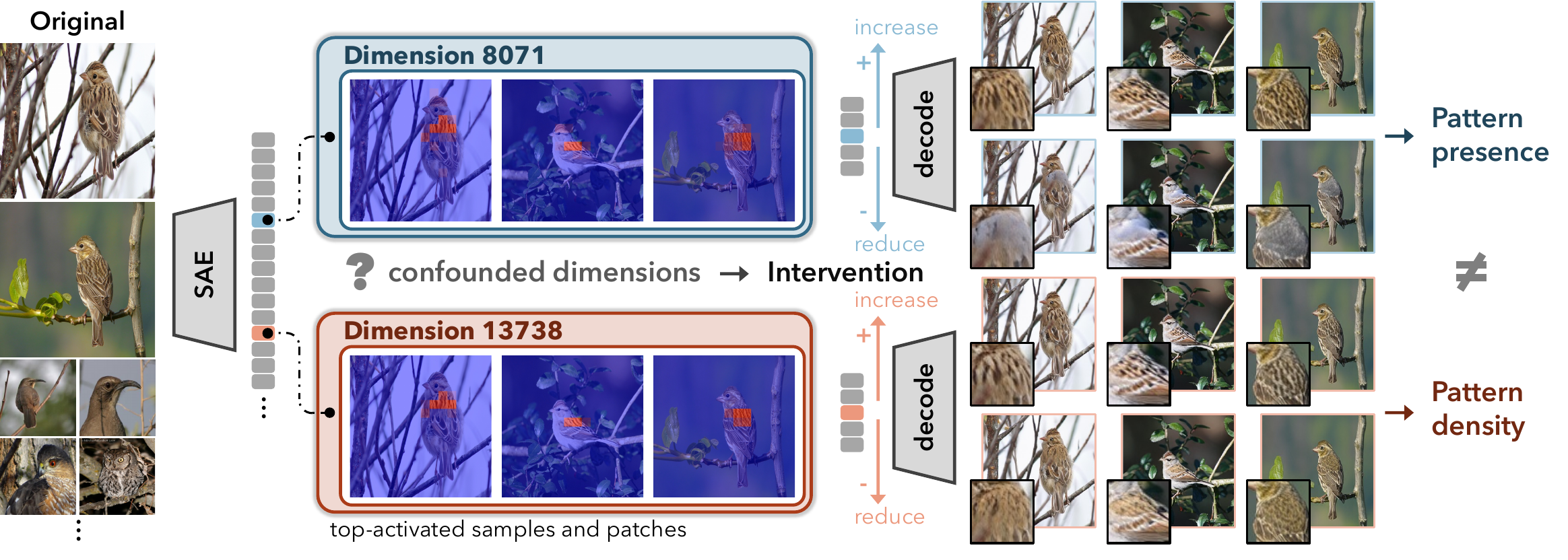}
    \caption{\small \textbf{Fine-grained categories expose confounded SAE dimensions}. Top-activated samples and pixel regions cannot effectively interpret the semantics they represent. We apply \textbf{dimensional intervention} on patch embeddings and decode them back to images. Comparing the visual changes from reducing or increasing the corresponding activation more clearly reveals the patterns or semantics represented by various SAE dimensions.}
    \label{fig:teaser}
\end{figure}

\begin{abstract}
Sparse autoencoders (SAEs) decompose visual representations into individual features, offering a promising basis for explaining model behavior. Interpreting these features typically relies on images or pixel regions that most strongly activate them. However, this evidence becomes ambiguous when multiple features with different semantics frequently activate in the same locations, such as texture and color in the same patterns. We propose \textbf{Reconstructive SAE Intervention} (\textbf{\method}), a framework that shifts feature explanation from passive observation to active visual reconstruction. By intervening on representations along SAE decoder directions and reconstructing the visual output, we provide direct evidence of a feature's semantic meaning. We demonstrate that our method successfully disambiguates feature meanings across multiple datasets and vision encoders. Furthermore, our framework grounds \class SAE interventions in a sparse set of patch SAE decoder directions. Beyond stable target response, this attribution also exposes the compositional relationship between elementary visual patterns and high-level semantics. Our results advocate for intervention reconstruction as an essential complementary signal for explaining neural networks.
\footnote{Demo and code available at \href{https://imageomics.github.io/RecSAEI}{imageomics.github.io/RecSAEI}.}
\end{abstract}

\section{Introduction}

Vision transformers process visual evidence across patch and \class tokens to produce downstream predictions~\citep{dosovitskiy2021vit}. Identifying the precise evidence that supports these predictions, however, is difficult within dense representations~\citep{elhage2022toy,bau2017network}. Sparse autoencoders (SAEs) address this challenge by decomposing dense representations into individual sparse features, each corresponding to a dimension of the sparse representation~\citep{shu2025survey,huben2024sparse,fry2024towards,daujotas2024case,daujotas2024interpreting}. Understanding what these individual features represent is then essential for building meaningful explanations of model behavior.

For vision models, a common practice to interpret an SAE feature is to retrieve and visualize the components (either regions within a given image or images from a given dataset) that maximally activate it~\citep{rao2024discover,lim2025sparse,stevens2025interpretable}.
Such activation-prioritized visualizations suggest possible interpretations, but several visual properties may consistently co-occur, leaving an individual feature's meaning ambiguous. This problem is exacerbated in fine-grained scenarios~\citep{wei2021fine}: as shown in \autoref{fig:teaser}, two SAE dimensions can activate on nearly identical images and regions while representing different properties. Their shared activation patterns reveal where relevant evidence occurs, yet leave unclear which aspect of that evidence each feature encodes. 
Intervention approaches offer additional evidence by measuring how modifying a feature changes downstream predictions~\citep{chatzoudis2025visual,goyal2019cace,zhou2018revisiting,abid2022meaningfully,bau2020understanding}. Yet a change in a category prediction alone does not identify the visual property associated with the intervened feature. Resolving this ambiguity requires making the visual consequences of feature intervention directly observable.

We propose \textbf{Reconstructive SAE Intervention}, or \textbf{\method}, which couples \textbf{sparse-feature intervention} with \textbf{reconstruction} of the modified representation into an image.
We first apply this approach to SAEs trained on patch tokens. Each target feature defines a direction in the patch representation through its SAE decoder vector. We modify the original patch embeddings along this direction and reconstruct the resulting image using a representation autoencoder (RAE)~\citep{zheng2026diffusion,singh2026improved}.
Comparing reconstructions across intervention strengths \textbf{reveals the visual changes associated with the target feature}. As shown in \autoref{fig:teaser}, these changes distinguish the pattern's presence and density, despite their similar high-activation examples.

Patch-level intervention, however, often does not directly connect to downstream predictions.
Therefore, we also need to interpret \class SAE features, which integrate information from patch representations and contribute more closely to model behavior. Yet the RAE decoder reconstructs images from patch representations and cannot directly decode the \class token.
We therefore connect the target \class SAE feature \textbf{back to the patch SAE features through gradient attribution}. We identify patch features that are both active and influential for the target, then combine their SAE decoder vectors into an intervention direction. Modifying patch representations along this direction changes the target \class activation and enables reconstructing the associated visual change. 

Compared with direct steering along a dense gradient, \method produces \textbf{more stable visual change and consistent responses} in the target \class feature (\S\ref{sec:cls-intervention-quantitative}). The sparse attribution also \textbf{identifies the patch SAE features} that support each target \class feature (\S\ref{sec:decompose-cls-to-patch}). Our analysis shows that these supporting sparse components may have distinct activation patterns and intervention effects. Their joint contribution nevertheless supports a stable and coherent higher-level response.
More broadly, our results motivate shifting feature explanation \textit{from activation patterns alone to examining the visual changes they produce under intervention}. Such intervention-based explanations provide more precise evidence for feature semantics and trace how patch-level patterns support image-level concepts, opening a path toward richer mechanistic analyses of vision models. 

\section{Related Work}
\paragraph{Sparse Concepts in Visual Representations}
Understanding concepts encoded in visual representations helps reveal the evidence underlying model decisions. Existing methods relate representations to concepts through individual units~\citep{bau2017network}, predefined concept directions~\citep{kim2018interpretability}, concept bottlenecks~\citep{koh2020concept}, clustering~\citep{ghorbani2019towards}, or factorization~\citep{zhang2021invertible,fel2023craft}. Sparse dictionary learning provides a complementary direction by representing activations with a small number of elements from an overcomplete dictionary~\citep{olshausen1996emergence,aharon2006ksvd,donhauser2025towards}. Sparse autoencoders (SAEs) extend this direction with an encoder-decoder reconstruction objective~\citep{fry2024towards}. In vision, SAEs have been applied to image-level embeddings~\citep{fry2024towards,daujotas2024case,daujotas2024interpreting,rao2024discover} and multimodal representations~\citep{shen2026vl}. More recent work applies SAEs to patch embeddings to expose spatially localized features~\citep{lim2025sparse,stevens2025interpretable}, evaluates their monosemanticity~\citep{pach2026sparse}, and identifies encoded concepts such as visual biases~\citep{choi2026conceptscope}. 

A central challenge for concept-based methods is establishing what an individual concept represents. Existing SAE studies commonly interpret each dimension from its highest-activating examples~\citep{zhang2025large,pahuja2026automatic}. \citet{ferrando2026language} use language models to summarize the effects of steering sparse features in visual embeddings, but rely on the aligned multimodal space of language models. Other work combines sparse features with explicitly interpretable concept spaces to improve robustness of interpretation~\citep{kulkarni2026interpretable}. Together, these methods provide rich descriptions of sparse features, but establishing their precise visual semantics remains challenging.

\paragraph{Intervention of visual concepts}
Many visual interpretability methods infer relevant evidence from activations, gradients, or input perturbations~\citep{selvaraju2017grad,wang2020score,chefer2021transformer,fong2017interpretable,goyal2019counterfactual,zhang2025finer,chowdhury2025prompt}.
Intervening further confirms the functions of that evidence by revealing changes in visual effects or predictions. Concept bottleneck models allow predefined concepts to be manipulated and propagate their effects to the output~\citep{koh2020concept,shin2023closer,steinmann2024learning,abid2022meaningfully}.
However, they cannot reveal the semantics of internal model representations. Prior work has also manipulated generative models for semantic image editing~\citep{goyal2019cace,shen2020interface,harkonen2020ganspace,shen2021sefa}. Recent diffusion methods realize fine-grained generation control~\citep{hertz2023prompt,tumanyan2023plug,brack2023sega,gandikota2024concept}. However, plugging concepts from discriminative models into generative models often needs dedicated training for each representation type. The iterative denoising process also adds uncertainty in the faithfulness of the intervened feature.


SAE provides another form of direct latent intervention. Manipulating the decomposed sparse features can mitigate object hallucination~\citep{park2026save}, improve zero-shot classification~\citep{chatzoudis2025visual}, and steer diffusion generation~\citep{kim2025concept}. \citet{carlyn2026simple} further use sparse latent interventions to produce localized counterfactual edits. These methods primarily use intervention to control model behavior or generation. \citet{kulkarni2026interpretable} steer SAE dimensions to explain different features, but rely on language models. We instead use reconstruction to recover the visual change and use it as evidence for disambiguating the semantics of the sparse feature.
\section{Methodology}

Assume an SAE model trained on the internal representations of a pretrained vision encoder, \textit{e.g.}, DINOv3~\citep{simeoni2026dinov3}.
Interpretability methods can identify the correlation between specific sparse dimensions and downstream predictions~\citep{steinmann2024learning,chatzoudis2025visual}.
We then need to understand the semantics of the visual evidence to fully explain model behavior.
For a sparse dimension $k$, a common way to infer its semantics is to inspect examples and spatial locations with large activation~\citep{zhang2025large,pahuja2026automatic}. Such evidence describes the inputs associated with a feature, but does not isolate the exact semantics. In fine-grained representations, in particular, the same image pixels may contain several co-occurring properties, such as the presence/absence and density of a pattern that can be used to differentiate categories or variants. High-activating examples alone cannot reliably distinguish such ambiguity.

Intervention provides complementary functional evidence by directly modifying a feature and observing its downstream effect~\citep{shin2023closer,steinmann2024learning}. However, the connection between a sparse dimension $k$ and prediction still does not specify the represented property. To this end, we propose \textbf{Reconstructive SAE Intervention} (\method), which couples sparse-feature intervention with \emph{reconstruction} to decode the intervention back into image space and inspect the corresponding visual change. Specifically, we use a pretrained RAEv2 representation decoder~\citep{singh2026improved}. Given patch representations from the corresponding vision encoder, the decoder reconstructs an image. Decoding intervened embeddings thereby exposes the visual consequence of the intervention. In this way, representation reconstruction enables translation between feature intervention and appearance changes, offering a mechanism to probe sparse visual features beyond activation patterns. The question, then, is \textit{how we can derive a reasonable intervention that can be reliably decoded}. We start with intervening on a single patch SAE dimension.

\begin{figure}
    \centering
    \includegraphics[width=\linewidth]{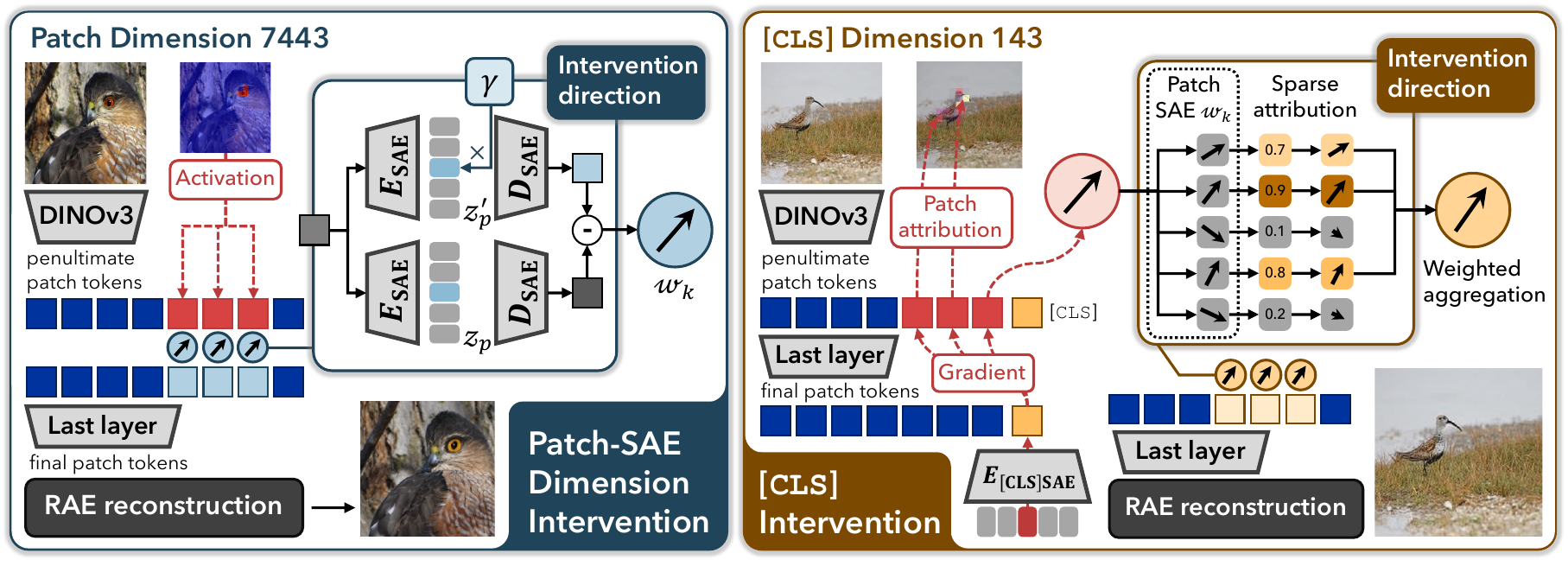}
    \vspace{-14pt}
    \caption{\small \textbf{Overview of \method for patch and \class SAE dimensions}. \textbf{Left}: For a patch-SAE feature, we identify highly activated spatial patches and intervene on the corresponding sparse dimension along the decoder direction \(\vw_k\). \textbf{Right}: For a \class feature, we first attribute its activation to patch tokens through backpropagation. After decomposing the patch-level gradient into patch-SAE dimensions, we aggregate top-$K$ decoder directions $\vw_k$ to obtain the intervention direction. We apply the intervention to attributed patches and decode changes through RAE reconstruction.}
    \label{fig:pipeline}
    \vspace{-6pt}
\end{figure}

\subsection{Intervention on a Patch SAE Dimension}

Let $\vh_p\in \mathbb{R}^d$ denote the encoded representation of a spatial patch token $p$, $\vz_p=E_{\text{SAE}}(\vh_p)$ its sparse feature activations, and $D_\text{SAE}(\vz_p)=\mW\vz_p+\vb$ its SAE reconstruction. We intervene on the sparse dimension $k$ by scaling its activation as $z_{p,k}^{'}=\gamma z_{p,k}$, with the other dimensions unchanged. A naive approach would decode the modified sparse representation directly through $\vh_p^{'}=D_\text{SAE}(\vz_p^{'})$. However, SAE reconstruction may introduce errors across all dimensions that conflate the intended feature intervention. Therefore, as shown in the left panel of \autoref{fig:pipeline}, we retain the original patch tokens and add only the change induced by the selected feature:
\[
\vh^{'}_p=\vh_p+\left[D_\text{SAE}(\vz^{'}_p)-D_\text{SAE}(\vz_p)\right]=\vh_p+\mW(\vz^{'}_p-\vz_p)=\vh_p+(\gamma-1)z_{p,k}\vw_k,
\]
where $\vw_k$, the $k$-th column of $\mW$, is the decoder direction associated with sparse dimension $k$. The intervention inherently moves $\vh_p$ along the feature's direction in representation space. This formulation recovers the original representation when $\gamma=1$. It also enables further extrapolation in reducing the feature activation below $0$.

Applying intervention across all spatial patches may induce unintended changes in image regions that contribute less to the feature's representation. Therefore, we restrict each intervention to its dominant spatial support. Specifically, we rank patches by feature activation and retain the smallest set that accounts for at least $80\%$ of the total activation mass, with at least four patches. The excluded patches are left unchanged. We apply the interventions on the input patch embeddings of the last layer and propagate them through the remaining encoder computation, after which they are decoded into full images using the RAE decoder~\citep{singh2026improved}.

\subsection{Intervention on \class Features}
Whereas patch-level SAE features are defined on individual spatial token representations, \class features describe more complex information, integrated from preceding patch tokens~\citep{dosovitskiy2021vit}. Intervening on these features reveals higher-level semantics and connects them to downstream predictions. However, the RAE decoder reconstructs images from patch representations. Directly modifying the final \class token therefore does not produce a corresponding image intervention. To this end, we propose to attribute the target \class feature back to the patch tokens and transform this signal into a sparse combination of patch-SAE decoder directions. Steering the selected patch representations along the aggregated direction and propagating the intervention through transformer blocks will alter the target \class feature and produce a corresponding visual reconstruction. 

Assume a target SAE trained on final-layer \class representations and a source SAE trained on penultimate-layer patch representations. Let \(y_c\) denote the activation of target \class SAE feature \(c\). We backpropagate the activation to derive patch-level gradients $\vg_p=\nabla_{\vh_p}y_c$, and localize the intervention by selecting the smallest set of patches $\mathcal{P}$ accounting for at least $80\%$ of the total gradient-norm mass. We can use the gradient as a \textit{dense} steering baseline on the selected patches to accordingly change the target \class SAE feature.

\paragraph{Sparse attribution}
While dense gradients directly optimize the response of a target \class dimension, we investigate whether a small set of supporting patch-level features can also control this response, providing a structured basis for examining the visual semantics of \class representations.
We therefore further express the target dimension’s sensitivity in the source SAE dictionary. Let \(z_{p,k}\) be the source activation of feature \(k\) at patch \(p\), and \(\vw_k\in\mathbb{R}^{d}\) its decoder vector. The directional sensitivity of the target to this feature is $q_{p,k}=\vg_p^\top \vw_k$. To identify features that are both present in the image and influential for the target, we aggregate absolute activation-weighted sensitivities for each source dimension over the selected patches:
\[
\sigma_k=\sum_{p\in\mathcal{P}}\,\bigl|z_{p,k}q_{p,k}\bigr|.
\]We retain the \(K\) highest-scoring features with nonzero attribution, forming a set \(\mathcal S_K\). If fewer than \(K\) features have nonzero scores, we retain only those features. Absolute values prevent positive and negative contributions at different locations from cancelling during feature selection.

\paragraph{Steering aggregation}
Using the selected source features, we construct a spatially varying steering direction by combining their decoder vectors with signed directional sensitivity weights:
\[
\vv_p=\sum_{k\in\mathcal S_K}q_{p,k}\vw_k,
\qquad
\vu_p=\frac{\vv_p}{\sqrt{\sum_{p\in\mathcal{P}}\|\vv_p\|^2_2}}.
\]
Here, the signed sensitivity $q_{p,k}$ determines both the magnitude and direction of each source feature's contribution. The normalization fixes the overall steering-direction norm across the selected patches $\mathcal{P}$. We then apply the intervention to the selected patches:
\[
\vh'_p = \vh_p + s\,\|\mH\|_{\mathrm F}\,\vu_p,
\]where \(s\) controls the intervention strength and $\|\mH\|_{\mathrm F}$ is the Frobenius norm of all patch embeddings. Scaling by this aggregate norm makes $s$ a relative intervention magnitude and ensures that the dense baseline and aggregated steering have the same intervention budget over the selected patches. We pass the edited patches through the final encoder block and reconstruct the image using the RAE decoder. The semantics revealed by observing visual changes after reconstruction can then be incorporated with other evidence to explain model behavior more accurately.
\section{Experiments}

\subsection{Intervention as Explanation}\label{sec:intervention-quantitative}

We first evaluate whether the intervened reconstruction provides meaningful information for explaining patch sparse features. We perform the study on four datasets: NABirds~\citep{van2015nabirds}, CUB~\citep{wah2011cub}, Cars~\citep{krause2013cars}, and ImageNet~\citep{deng2009imagenet}.
In fine-grained scenarios, sparse features represent subtler semantics than in coarse-grained datasets. CLIP~\citep{radford2021clip} or GroundingDINO~\citep{liu2024grounding} models may not accurately identify fine-grained organisms. Therefore, IoU or CLIP score adopted in previous literature~\citep{xu2025deciphering,zhang2025large} cannot faithfully reflect explanation quality (results in \S\ref{sec:app-clip-iou}). Moreover, many features are hard for humans to describe and difficult to match to a predefined dictionary.

To this end, we propose to predict feature activation on held-out natural images. Specifically, for each sparse feature, we construct its explanation using the training set. Given two test images and the explanation, an evaluator predicts which image has a higher activation for this feature. 

\paragraph{Experiment setting}
We build a paired benchmark for this evaluation.
First, we exclude dense (activated on more than $0.1\%$ of images) and dead dimensions. Within the remaining dimensions, we exclude dimensions with more than 20\% activation on the backgrounds. We use bounding-box labels from NABirds, CUB, and Cars to indicate foreground, and generate bounding boxes for ImageNet using GroundingDINO and the category name. From the resulting candidate pool, we randomly select $1{,}000$ dimensions to build the benchmark. For each dimension, we select five image pairs with high cosine similarity in the dense embedding space but exhibit a strong contrast in SAE activation. \S\ref{sec:app-pair-benchmark} and \autoref{tab:app-dimension-selection} present more details of the benchmark.
We use GPT-5.6-Terra~\citep{openai2026gpt56} to predict the sample with higher activation values and compare the following forms of explanations: (1) text explanations generated by observing the top examples, \textit{i.e.}, images with high activations in the target feature; (2) top examples; (3) top examples with intervention reconstructions; (4) top examples and bottom examples, \textit{i.e.}, images with low activations; (5) the combination of top examples, intervened reconstructions, and bottom examples. We provide the activation overlay along with top examples to indicate feature location (Ablation in \S\ref{sec:app-activation-form}). We include embeddings from DINOv3~\citep{simeoni2026dinov3} and SigLIP 2~\citep{tschannen2025siglip2}. Training and intervention details are included in \S\ref{sec:app-implementation}.

\begin{table}[t]
    \centering
    \small
    \caption{\small \textbf{Feature-explanation accuracy using different evidence}. Text: descriptions of the target feature. Top: images and regions with high activations. Bottom: images with low activations. Intervention: reconstructions after reducing and increasing the feature activation. Intervention reconstructions provide information complementary to both top and bottom examples to explain SAE features.}
    \label{tab:main-paired}
    \vspace{-4pt}
    \setlength{\tabcolsep}{4pt}
    \resizebox{\textwidth}{!}{
    \begin{tabular}{@{}ccccCCCCCCCC@{}}
    \toprule
        \multicolumn{4}{c}{\textbf{Explanation approach}} & \multicolumn{2}{c}{\textbf{NABirds}} & \multicolumn{2}{c}{\textbf{CUB}} & \multicolumn{2}{c}{\textbf{Cars}} & \multicolumn{2}{c}{\textbf{ImageNet}} \\
        Text & Top & Bottom & Intervention & \text{DINOv3} & \text{SigLIP 2} & \text{DINOv3} & \text{SigLIP 2} & \text{DINOv3} & \text{SigLIP 2} & \text{DINOv3} & \text{SigLIP 2} \\
    \midrule
        \checkmark & & & & 
        62.8 & 59.8 & 59.0 & 59.1 & 61.5 & 61.4 & 74.1 & 71.1 \\
        & \checkmark & & & 
        66.3 & 61.0 & 64.3 & 63.1 & 67.4 & 65.5 & 74.7 & 71.8 \\
        & \checkmark & & \checkmark &  
        68.2 & 63.8 & 65.4 & 64.8 & 68.4 & 66.0 & 71.4 & 72.3 \\
        & \checkmark & \checkmark & &
        73.0 & 67.9 & 70.3 & 69.7 & 73.2 & 69.4 & \mathbf{76.9} & 73.5 \\
        & \checkmark & \checkmark & \checkmark & 
        \mathbf{74.8} & \mathbf{69.2} & \mathbf{71.4} & \mathbf{71.1} & \mathbf{74.3} & \mathbf{71.0} & 76.2 & \mathbf{73.9} \\
    \bottomrule
    \end{tabular}
    }
\end{table}

\paragraph{Results}
We present the accuracy of predicting the sample with higher activations in \autoref{tab:main-paired}.
Noticeably, top-activating examples already outperform text-only explanations across all settings, even though the descriptions are produced by the same GPT-5.6-Terra model. Directly comparing the test images with examples facilitates more flexible and comprehensive reasoning. Bottom-activating examples can be treated as indirect intervention on the target dimension, which provides a substantial performance gain, suggesting that contrasting when a feature is present and absent is highly informative. However, bottom examples cannot directly disambiguate confounded features.

Intervention reconstruction (\method) contributes a further, largely complementary signal. When added to top and bottom examples, it improves explanation accuracy in seven of eight settings. The improvement is more significant in fine-grained datasets, including gains of \(1.8\) points on NABirds with DINOv3 and \(1.6\) points on Cars with SigLIP 2. ImageNet contains coarser-grained categories, making it easier to identify the higher-activating test image. Nevertheless, intervention reconstruction still improves the explanation accuracy for SigLIP 2.
Overall, the results suggest that feature activation and intervention characterize different aspects of a sparse dimension: activation indicates the visual contexts associated with the feature, while intervention helps isolate the visual property controlled by that feature.
Observing how a feature changes visual appearance provides semantic evidence that its activation pattern alone cannot recover.
We also evaluate text descriptions produced with different visual components in \S\ref{sec:app-text-explanation-comparison} to avoid bias from visual similarity.

\begin{figure}[t]
\centering
    \includegraphics[width=0.9\textwidth]{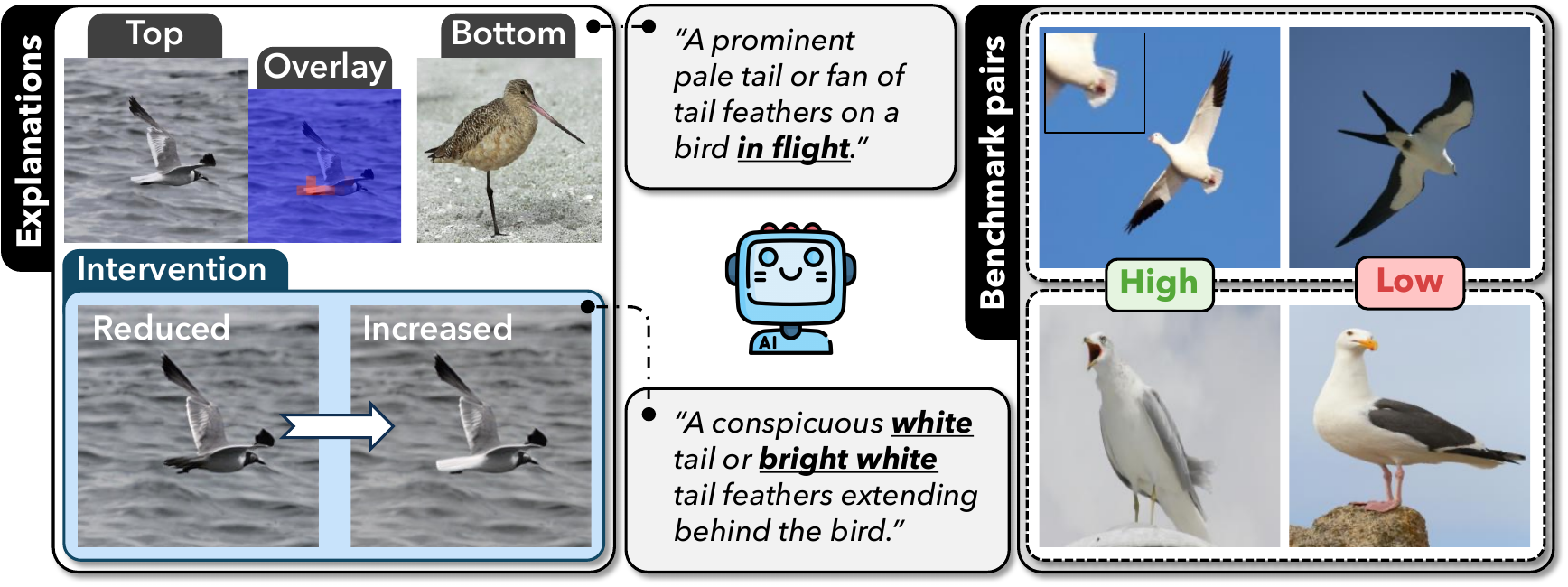}
    \vspace{-4pt}
    \caption{\small \textbf{Qualitative example of intervention-assisted feature explanation}. Top and bottom examples alone might bias the benchmark prediction toward pose, prominence, or ``in-flight'' status. Intervention reconstruction isolates the more specific change toward bright white tail feathers, leading to a more precise feature interpretation.}
    \label{fig:explanation-example}
    \vspace{-6pt}
\end{figure}

\paragraph{Qualitative analysis}
\autoref{fig:explanation-example} illustrates the complementary role of intervention reconstruction beyond activation-based evidence. From the top and bottom examples alone, the feature is broadly associated with pale tail feathers, but it may also relate to tail pose, tail prominence, or in-flight status. Intervention reconstruction isolates a more specific visual change: increasing the feature brightens the tail toward white. This additional evidence produces a more precise interpretation and helps distinguish benchmark pairs whose activation patterns are visually similar.

\subsection{\class Intervention Evaluation}\label{sec:cls-intervention-quantitative}

We then evaluate the effectiveness of \method on \class sparse features, and more specifically, the necessity of sparse attribution and steering aggregation. 

\paragraph{Experiment setting}
We evaluate \class sparse feature interventions on NABirds using a frozen DINOv3 encoder and RAEv2 decoder, with separate SAEs trained on final-layer \class tokens and penultimate-layer patch tokens. We sample $128$ \class features with five validation images per feature. For each image–feature pair, we compare the proposed sparse steering with a dense baseline that directly uses the patch-level gradient direction for steering. Both settings apply interventions on the same selected patches and share the same intervention budget, varying the number of contributing SAE features $K$ and intervention strength. We use the unedited reconstruction as a reference, re-encode the edited image, and measure round-trip activation changes on the target \class dimension, as well as off-target changes.

\begin{figure}[t]
    \centering
\begin{subfigure}{0.32\textwidth}
    \includegraphics[width=\textwidth]{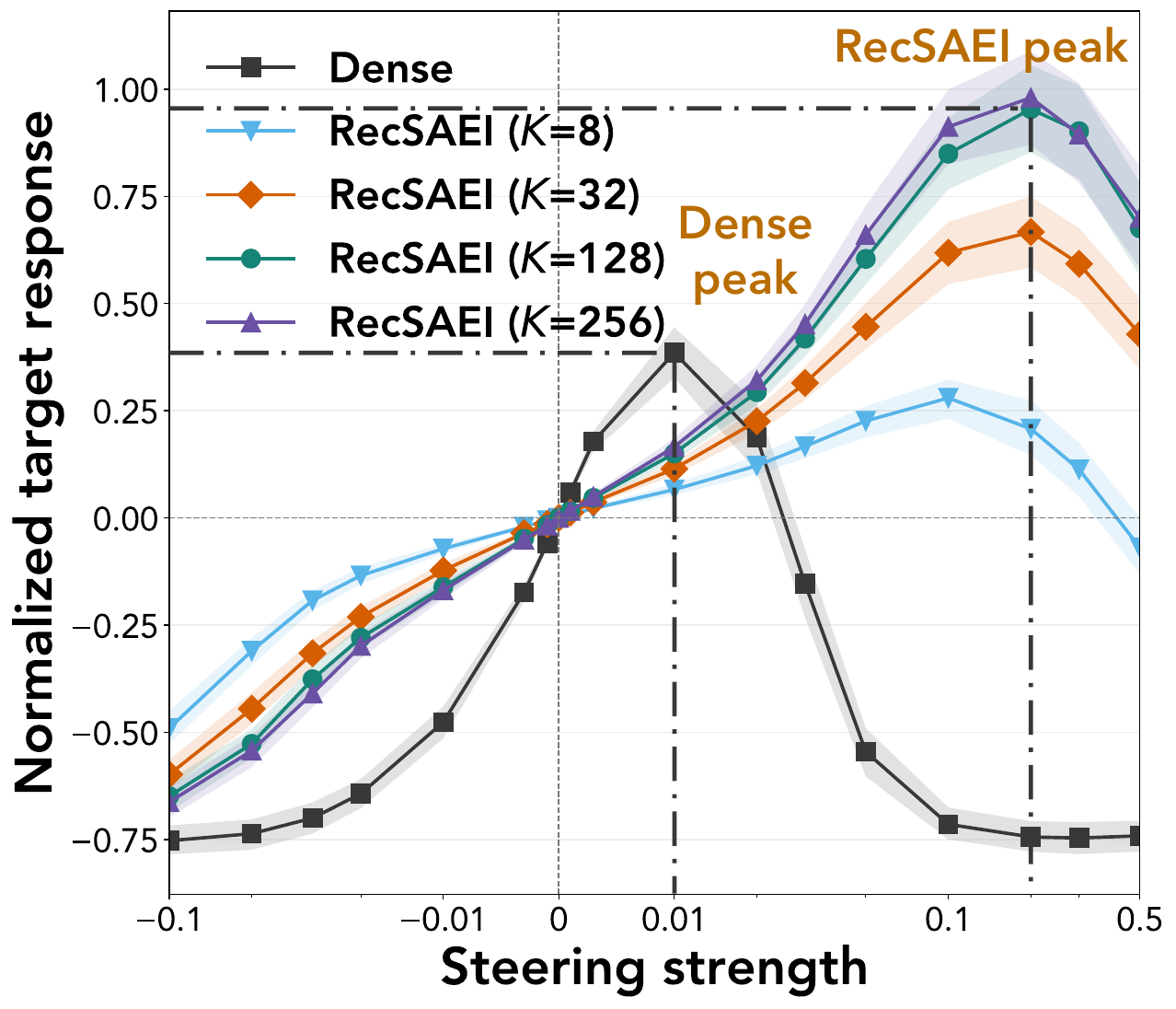}
\vspace{-14pt}
    \caption{}
    \label{fig:target-response}
\end{subfigure}
\hfill
\begin{subfigure}{0.32\textwidth}
    \includegraphics[width=\textwidth]{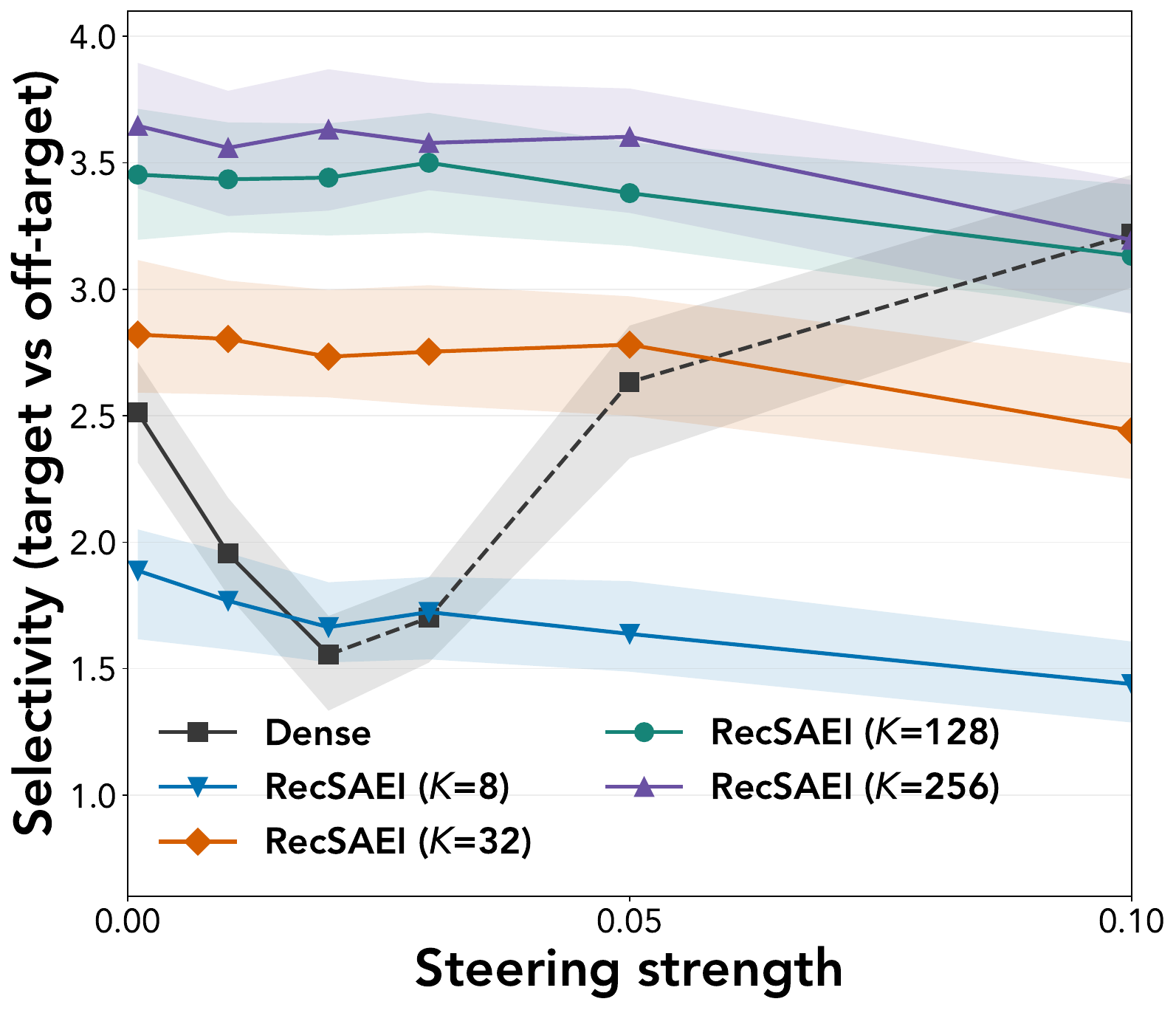}
\vspace{-14pt}
    \caption{}
    \label{fig:selectivity-strength}
\end{subfigure}
\hfill
\begin{subfigure}{0.32\textwidth}
    \includegraphics[width=\textwidth]{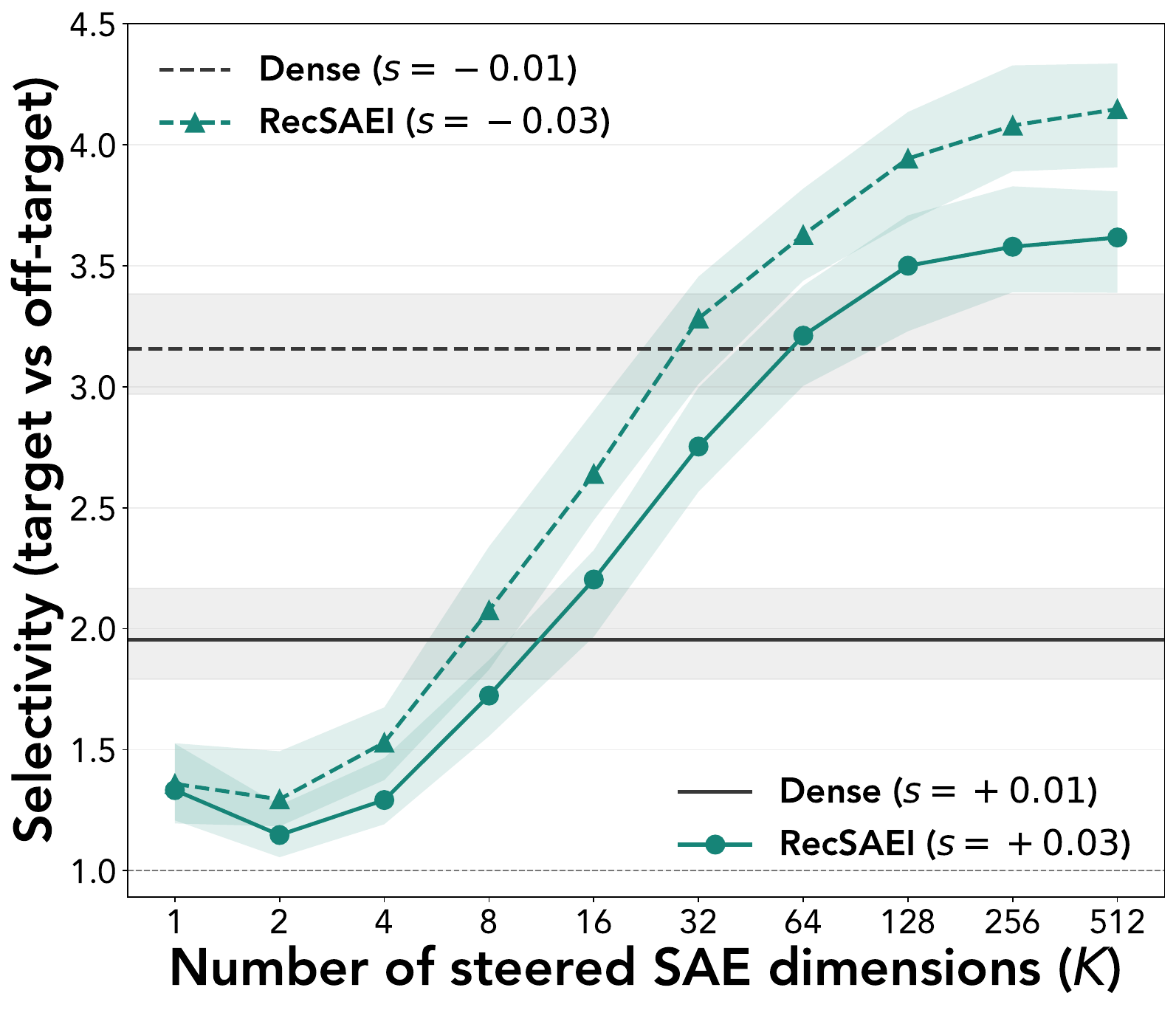}
\vspace{-14pt}
    \caption{}
    \label{fig:selectivity-k}
\end{subfigure}
\vspace{-6pt}
\caption{\small\textbf{\method provides more consistent and selective control} of \class SAE features than dense steering. (a) Target \class response after re-encoding the reconstructed image across intervention strengths. Dense steering leads to image collapse under a small steering strength, while \method saturates later and provides a higher peak response. (b) Target-to-off-target selectivity for different numbers of attributed patch-SAE features. The dense curve with $s>0.01$ is dashed, as the reconstruction already collapses. (c) Target-to-off-target selectivity against the number of steered dimensions \(K\). We select $s=\pm 0.01$ for dense and $\pm 0.03$ for \method to roughly match the target response, as shown in \autoref{fig:target-response}.}
\vspace{-6pt}
\end{figure}

\begin{figure}[t]
    \includegraphics[width=\textwidth]{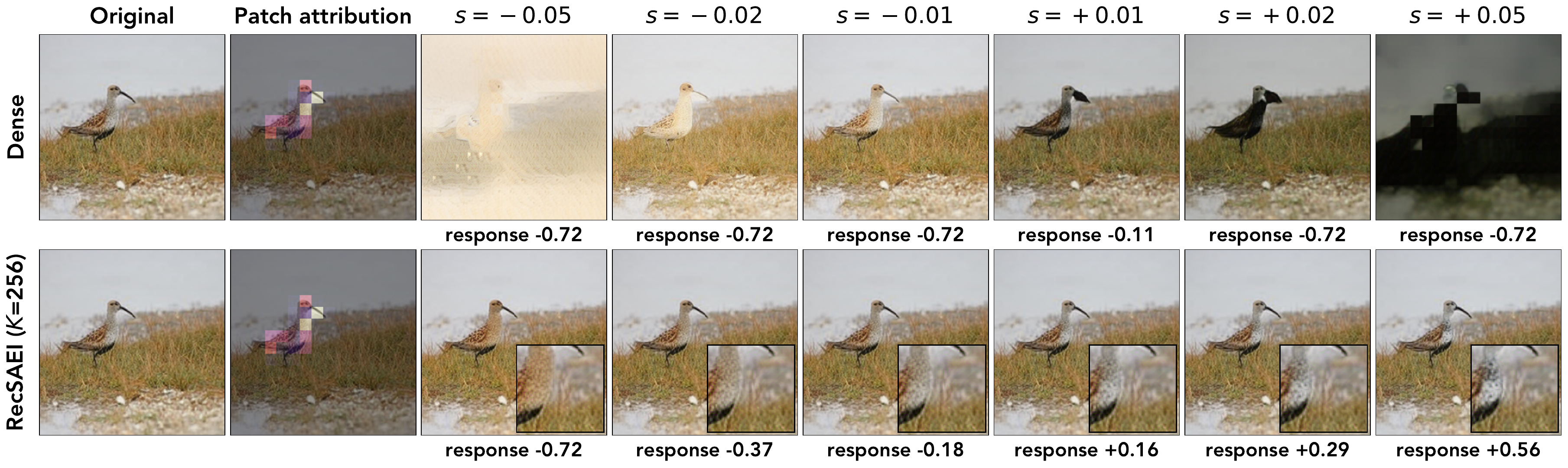}
    \vspace{-14pt}
    \caption{\small\textbf{Qualitative comparison of dense and sparse \class steering}. Dense steering induces large global appearance changes and an inconsistent target response, whereas \method gradually changes the bird's neck color from brown to grey with black spots, with a consistent \class response across intervention strengths. The original column is a direct RAE reconstruction without intervention.}
    \label{fig:reconstruction-comparison}
    \vspace{-4pt}
\end{figure}

\paragraph{Results}
\method provides substantially more consistent control over target \class features than dense steering. As shown in \autoref{fig:target-response}, dense steering exhibits a narrow effective range. Increasing the intervention strength $s$ above $0.01$ leads to degraded response. In contrast, aggregating attributed patch SAE directions provides an approximately monotonic target response until $s>0.2$ and a higher peak response. \method also improves intervention selectivity. \autoref{fig:selectivity-strength} shows that when more than $32$ patch dimensions are aggregated, the target feature changes more strongly relative to off-target \class features. Selectivity decreases slightly under larger intervention strengths, but much more slowly than dense steering. \autoref{fig:selectivity-k} further shows that selectivity improves as aggregation involves more patch SAE dimensions, and $32$ patch dimensions stably exceeds the corresponding dense-steering baseline. Results with more steering strengths are available in \S\ref{sec:app-more-strengths}.

\begin{figure}[t]
    \includegraphics[width=\textwidth]{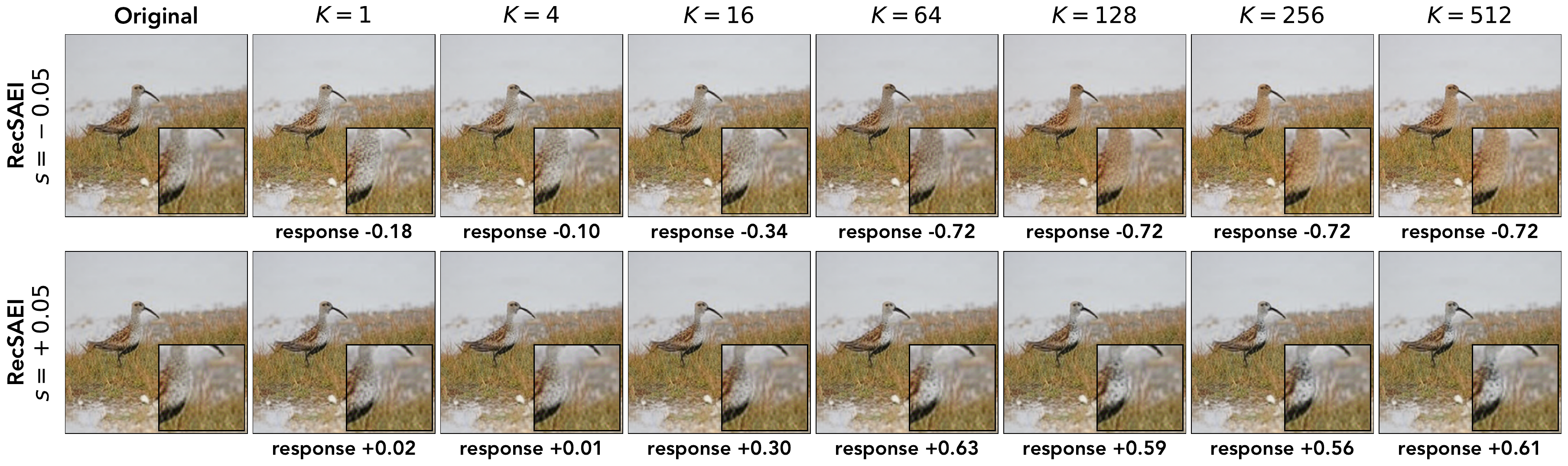}
    \vspace{-14pt}
    \caption{\small\textbf{Effect of the number of steered patch-SAE features $K$}. A small $K$ involves a limited number of patch-SAE features and produces only slight visual changes. Increasing \(K\) makes changes more significant in both appearance and target response while preserving the overall image appearance. The intervention converges once the major contributing features are included ($K\ge 128$), validated by the target response. The original column is a direct RAE reconstruction without intervention.}
    \label{fig:reconstruction-varyk}
    \vspace{-6pt}
\end{figure}

\paragraph{Qualitative analysis}
\autoref{fig:reconstruction-comparison} reveals the corresponding difference in image space. Dense steering quickly introduces large global changes in illumination and appearance. The collapsed reconstruction also fails to increase the target \class feature. In contrast, \method produces gradual visual changes concentrated around the attributed regions (from brown color to grey with black spots) and a consistent response in the target \class activation. \autoref{fig:reconstruction-varyk} shows that increasing the number of involved patch-SAE dimensions $K$ produces more significant visual changes while preserving the overall image structure. At the same time, the induced target response change also becomes more stable. More intervention examples are in \S\ref{sec:app-qualitative}.

\subsection{Analysis and Discussions}

\begin{figure}[t]
    \includegraphics[width=\textwidth]{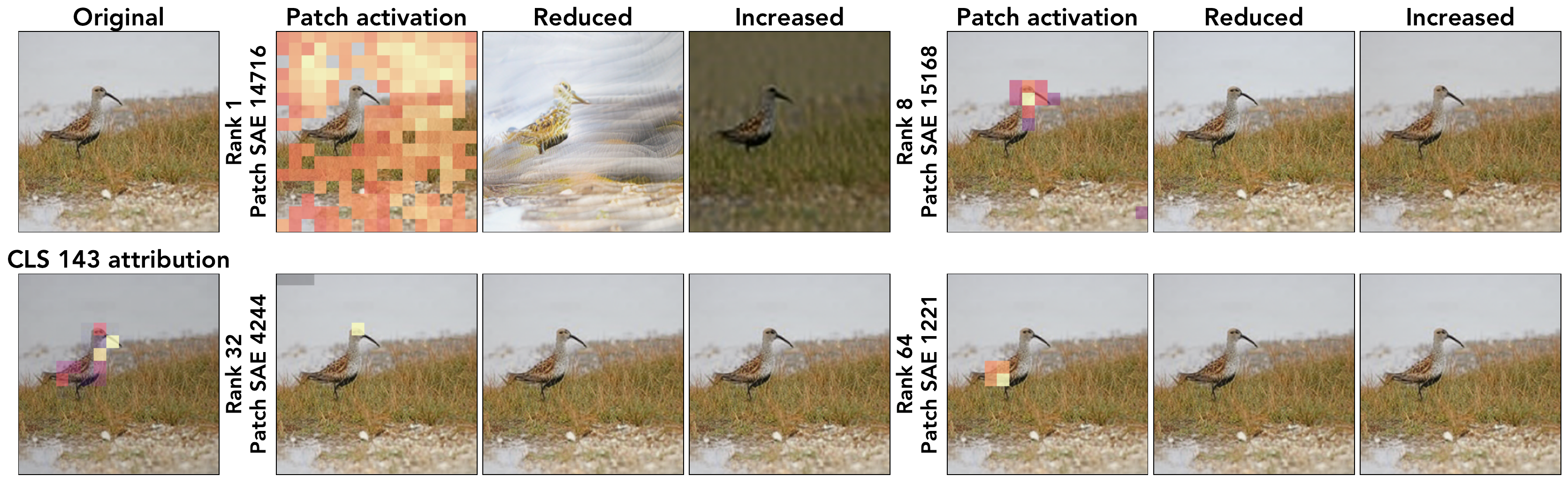}
    \vspace{-16pt}
    \caption{\small\textbf{Patch-SAE decomposition of a \class SAE feature}. Highly attributed patch SAE dimensions localize to different image regions and induce distinct visual changes under intervention, revealing the local sparse components that jointly support the global \class feature. The original column is a direct RAE reconstruction without intervention. We apply interventions to patches with the feature activated instead of attributed patches.}
    \label{fig:component-interventions}
    \vspace{-8pt}
\end{figure}

\paragraph{Decomposing \class to Patch Sparse Features}\label{sec:decompose-cls-to-patch}
Beyond enabling stable steering, our formulation exposes the patch-level components underlying a target \class SAE feature. \autoref{fig:component-interventions} shows several of the highest-attributed patch SAE dimensions for the same target \class feature. These dimensions exhibit distinct activation patterns and intervention effects, capturing heterogeneous visual properties, such as object presence, color, and shape. 

The top-ranked dimension corresponds to the ``elsewhere'' concept, which is essential in indicating the object's presence~\citep{fel2026into}. This dimension produces significant visual changes when intervened on. However, \method only modifies the attributed patch tokens (shown in \autoref{fig:reconstruction-comparison}), which explains the marginal visual change in \autoref{fig:reconstruction-varyk}. At the same time, this dimension also leads to relatively poor selectivity when $K$ is small. With more dimensions involved, their aggregated direction becomes more meaningful and produces more targeted changes. We list detailed statistics of sparse dimension attribution in \S\ref{sec:app-cls-attribution-stat}.
The decomposition provides an interpretable account of the steering direction, connecting the response of a global \class feature to the local visual ingredients that contribute to it.  We present more attributed patch SAE dimensions and examples in \S\ref{sec:app-dimension-decomposition}.

\begin{figure}[t]
\centering
\begin{subfigure}{0.49\textwidth}
    \centering
    \includegraphics[width=\textwidth]{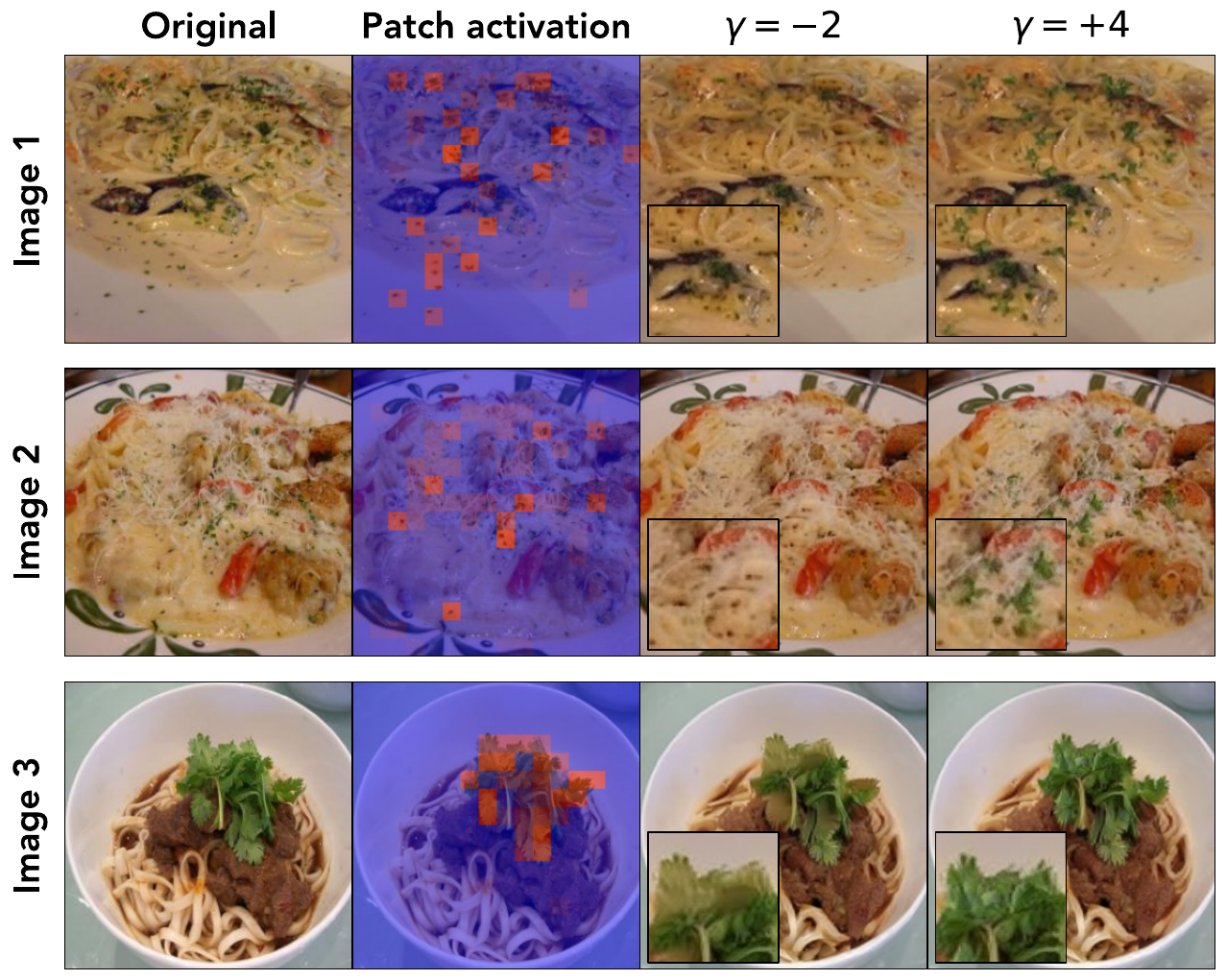}
    \vspace{-12pt}
    \caption{Patch SAE dimension $4443$ in ImageNet--SigLIP 2.}
\end{subfigure}
\hfill
\begin{subfigure}{0.49\textwidth}
    \centering
    \includegraphics[width=\textwidth]{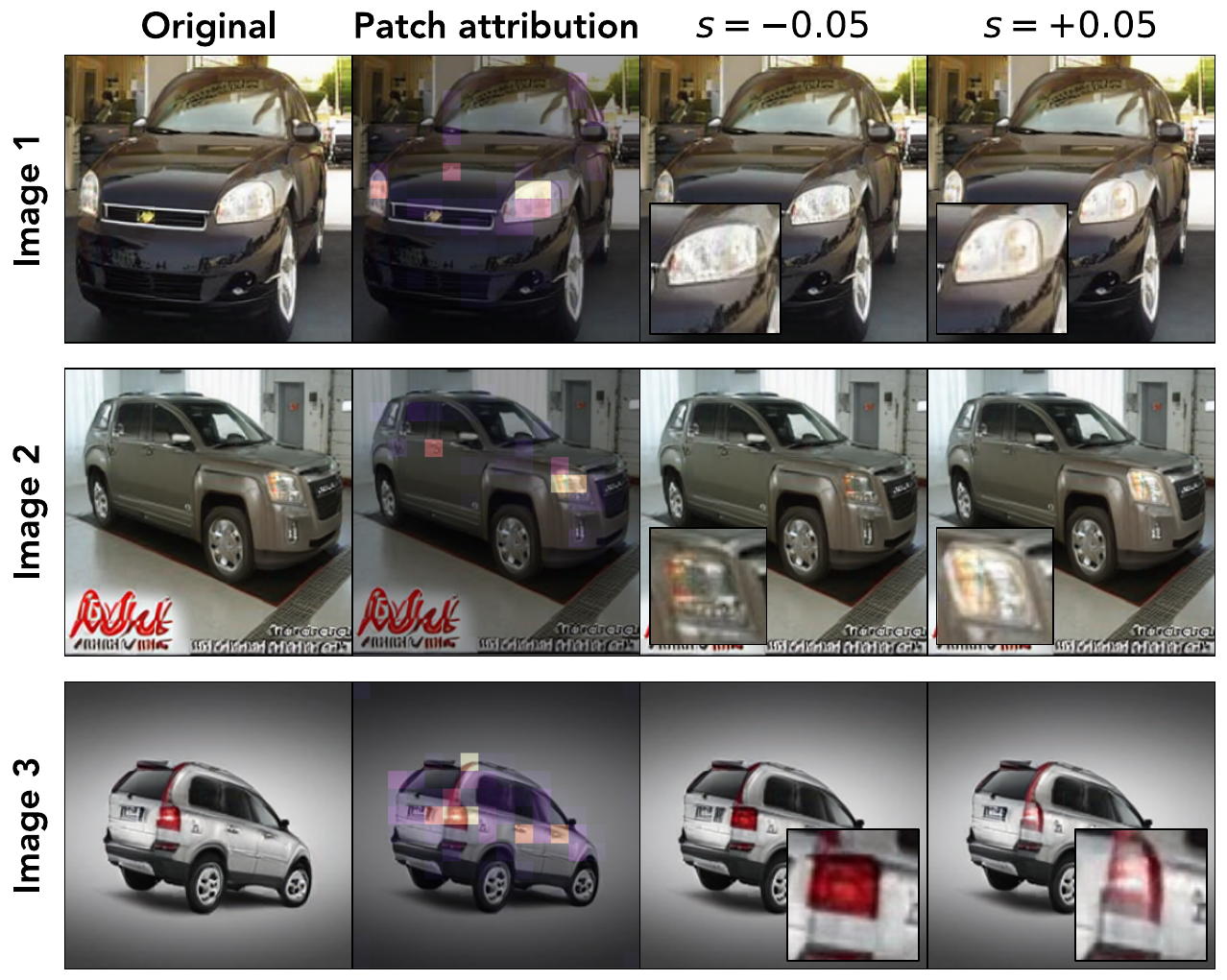}
    \vspace{-12pt}
    \caption{\class SAE dimension $412$ in Cars--DINOv3.}
\end{subfigure}
\vspace{-4pt}
\caption{\small\textbf{Cross-image consistency of intervention on (a) patch SAEs and (b) \class SAEs}. Patch-SAE intervention uses activation to select patches for intervention, while \class intervention uses gradient attribution. Both approaches identify relevant patterns, while negative and positive interventions consistently produce similar visual changes: patch dimension $4443$ yields more green herbs and \class dimension $412$ yields taller and brighter car headlights. The original columns are direct RAE reconstructions without intervention. $\gamma$ represents the scale of amplifying the target feature, and $s$ is the ratio of the original patch embedding norm.}\label{fig:steer-consistency}
\vspace{-8pt}
\end{figure}

\paragraph{Consistency of Intervention}
Beyond producing a stable response as intervention strength varies, the attributed sparse steering direction also generalizes consistently across different images activating the same patch and \class feature. \autoref{fig:steer-consistency} shows three examples each for a shared patch and \class SAE dimension. Although the spatial attribution varies with image content, applying the positive intervention generally creates more green herbs for patch SAE dimension $4443$ and taller headlights for \class dimension $412$. The corresponding reconstructions preserve the overall scene while introducing localized appearance changes around the activated or attributed regions. This cross-image consistency suggests that the intervention captures a reusable visual direction associated with the patch or \class feature.

\begin{wraptable}{r}{0.48\textwidth}
\vspace{-10pt}
\centering
\small
\captionsetup{type=table,hypcap=false}
    \captionof{table}{\small\textbf{Direction agreement of sparse-feature steering}. Steering generally changes the monitored feature in the intended direction, particularly for \class SAE intervention, but imperfect direction agreement and trajectory monotonicity indicate that feature control is not exact.}\label{tab:steering-direction}
\vspace{-4pt}
\resizebox{0.48\textwidth}{!}{
    \begin{tabular}{@{}l|CCC@{}}
    \toprule
        \textbf{Monitored} & \textbf{Direction} & \textbf{Trajectory} & \textbf{Strength} \\
        \textbf{target} & \textbf{correctness} & \textbf{monotonicity} & \textbf{correlation} \\
    \midrule
        Patch SAE  & 64.4\% & 63.1\% & 0.71 \\
        \class SAE & 87.0\% & 73.9\% & 0.93 \\
    \bottomrule
    \end{tabular}
}
\end{wraptable}

\paragraph{Intervention Agreement}
Beyond changes in visual appearance, we further check the agreement between the \method intervention and the response in target feature dimensions for both patch and \class SAEs. As shown in \autoref{tab:steering-direction}, for individual patch SAE features, direction correctness and trajectory monotonicity are \(64.4\%\) and \(63.1\%\), respectively, with a strength–response correlation of \(0.71\). Steering \class SAE features through attributed patch features is considerably more reliable, increasing these values to \(87.0\%\), \(73.9\%\), and \(0.93\). The results suggest that, by incorporating complementary signals from patch SAE features, \method attains more stable and coherent intervention on \class SAE features.
At the same time, the imperfect intervention agreement also reveals the limitations of existing SAE and RAE models, motivating us to incorporate corresponding agreement signals into SAE and RAE model training, which is an important future direction.
\section{Conclusion}
This paper introduces \method, an intervention-reconstruction framework for interpreting sparse visual features by decoding representation interventions into visual changes. For patch SAE features, steering along decoder directions reveals information complementary to activation-based evidence, helping disambiguate confounded features. For \class SAE features, we attribute global features to patch-level dimensions and aggregate top-contributing components for steering. The sparse formulation yields stable and interpretable interventions for otherwise undecodable \class SAE features.
More broadly, our results motivate rethinking the format of feature explanations. We should advocate for including \textit{how} representations change under intervention beyond static descriptions or high-activation examples to explain features comprehensively. We also need richer attribution approaches that trace how sparse features interact and compose across the network to faithfully relate visual evidence to downstream predictions.

\subsubsection*{Acknowledgments}
This research is supported in part by grants from the NSF Imageomics Institute (OAC 2118240) and the NSF and NSERC AI and Biodiversity Change Global Center (NSF Award No. 2330423 and NSERC Award No. 585136). The authors are grateful for the generous support of the computational resources from the Ohio Supercomputer Center. We also acknowledge the API support provided by the National Artificial Intelligence Research Resource (NAIRR) Pilot under Award No. 250205.

\bibliography{iclr2027_conference}
\bibliographystyle{iclr2027_conference}

\newpage
\appendix

\DoToC

\newpage

\section{Implementation Details}\label{sec:app-implementation}

\subsection{SAE Training}
We train SAE models for two architectures: \href{https://huggingface.co/facebook/dinov3-vitl16-pretrain-lvd1689m}{DINOv3-L/16}~\citep{simeoni2026dinov3} and \href{https://huggingface.co/google/siglip2-large-patch16-256}{SigLIP2-L/16}~\citep{tschannen2025siglip2}. Both of them use a ViT-large backbone and a patch size of $16$. Because SigLIP 2 does not have a conventional \class token, we only train \class SAEs for DINOv3.

\paragraph{Patch SAE}
We train the patch SAE model on the penultimate-layer (layer \(-2\)) activations extracted from images resized to \(256\times256\). It operates on the $256$ spatial patch tokens per image, using a $1{,}024$-dimensional input and a $16{,}384$-dimensional dictionary. We used a BatchTopK SAE~\citep{bussmann2024batchtopk} with \(k=32\), Adam optimization, and a learning rate of \(4\times10^{-4}\). The training budget was $100$ million token presentations for NABirds and ImageNet, and $25$ million for CUB and Cars. Before training, we shuffle the patch activations to mitigate location-induced correlations, followed by ordered inference on the training and validation sets.

\paragraph{\class SAE}
We train the \class SAE separately on the final-block (layer \(-1\)) \class token activations. It uses a $1{,}024$-dimensional input, a $4{,}096$-dimensional dictionary, and the same BatchTopK objective with \(k=32\), together with Adam and a learning rate of \(4\times10^{-4}\). Because each image only contributes one \class token, training uses a batch size of $512$ and two million token presentations. We set the dead-feature threshold to $100{,}000$ presentations. We keep both SAEs frozen during the subsequent attribution and steering experiments.

\subsection{SAE Statistics}
\begin{table}[h]
    \centering
    \small
    \caption{\textbf{Validation statistics of patch SAEs across datasets and visual backbones}. We report reconstruction MSE, \(L_1\) statistics, and the numbers of dead, almost-dead, and dense features for SAEs trained on DINOv3 and SigLIP 2 representations. }
    \label{tab:app-sae-stats}
    \begin{tabular}{@{}llCCCCC@{}}
    \toprule
        \textbf{Model} & \textbf{Dataset} & \textbf{MSE} & L_1 & \textbf{Dead} & \textbf{Almost dead} & \textbf{Dense} \\
    \midrule
        \multirow{4}{*}{DINOv3}
        & NABirds  & 62.15  & 2{,}579.18 & 0 & 0 & 432 \\
        & CUB      & 72.78  & 2{,}517.16 & 1 & 1 & 473 \\
        & Cars     & 75.49  & 2{,}577.52 & 2 & 2 & 363 \\
        & ImageNet & 114.62 & 2{,}813.61 & 55 & 166 & 407 \\
    \midrule
        \multirow{4}{*}{SigLIP 2}
        & NABirds  & 0.26  & 406.78 & 105 & 105 & 475 \\
        & CUB      & 0.31  & 358.01 & 78  & 78  & 544 \\
        & Cars     & 0.31  & 298.32 & 85  & 85  & 484 \\
        & ImageNet & 0.35  & 257.74 & 231 & 600 & 431 \\
    \bottomrule
    \end{tabular}
\end{table}

Table~\ref{tab:app-sae-stats} summarizes the validation statistics of the patch SAEs across datasets and backbones. We use activation frequency below $10^{-7}$ as the criterion for almost-dead dimensions and above $10^{-2}$ for dense dimensions.
The number of dense features remains relatively stable across datasets for both backbones. Relatively, ImageNet has more tokens and is thus less optimized. These results indicate that the learned sparse representations are broadly well utilized. 

\subsection{Steering Implementation}
We use \href{https://huggingface.co/nyu-visionx/RAEv2-models/tree/main/stage1/imagenet/dinov3l-k1}{dinov3l-k1} and \href{https://huggingface.co/nyu-visionx/RAEv2-models/tree/main/stage1/imagenet/siglip2-l-k1}{siglip2l-k1} RAEv2 decoders for DINOv3 and SigLIP 2, respectively.

\paragraph{Patch SAE}
For interventions on a single patch-SAE dimension, each image is first encoded to the SAE input layer and transformed into patch-level SAE codes \(\vz_{p}\). Given a target dimension \(k\), we modify only that coordinate by setting \(\vz'_{p,k}=\gamma \vz_{p,k}\), while leaving all other SAE dimensions unchanged. We then decode both the original and edited SAE codes and add the decoder difference, \(D_\text{SAE}(\vz'_p)-D_\text{SAE}(\vz_p)\), to the original dense patch representations, thereby avoiding errors from SAE reconstructions. The intervention is restricted to the spatial patches containing $80\%$ of the dimension’s positive activation mass, with at least four patches selected. We leave the \class and register tokens unmodified. We pass the patch embeddings through the remaining encoder and use the RAE decoder to map the intervened embeddings back to an image.
For the direction agreement experiment in \autoref{tab:steering-direction}, we re-encode the produced images through the original encoder and patch SAE. We evaluate suppression and amplification using scales \(\gamma\in\{0,0.5,1,2,4\}\), where \(\gamma=0\) removes the feature and \(\gamma>1\) amplifies it.

\paragraph{\class SAE}
We do not directly edit \class SAE dimensions, as the RAE decoder does not take \class embeddings as inputs. Instead, we treat each target \class SAE dimension as a scalar objective and intervene upstream on the penultimate-layer patch representations that influence it. For each target dimension, we compute its gradient with respect to the raw patch tokens $\vg_p=\nabla_{\vh_p}y_c$. As in patch-SAE intervention, we steer only the smallest set of patches containing $80\%$ of the gradient mass. The sparse attribution first decomposes gradients into patch-SAE dimensions by decoder-direction alignment $q_{p,k}=\vg_p^\top \vw_k$, which indicates their sensitivity to the target change. 
We rank patch-SAE dimensions by the product of dimension activation and sensitivity and select the top \(K\) dimensions for the subsequent intervention.
The aggregation step combines their decoder directions using the signed gradient weights and normalizes the directions across patches. Finally, the intervention acts similarly to the patch-SAE intervention and applies a steering strength toward the aggregated direction to the selected patch embeddings. Here, for better control, we define the intervention magnitude relative to the Frobenius norm of the original patch-token tensor. This way, dense steering and SAE-derived steering receive the same amount of input-level changes.

We propagate the updated patch tokens through the remaining encoder and RAE decoder, keeping \class and register tokens unchanged. We then re-encode the edited image and measure changes in the target \class SAE feature’s dense and sparse activations \(h\) and \(f\), together with off-target changes and selectivity. The intervention controls a \class SAE feature indirectly through an attributed subspace of patch-SAE decoder directions.

\subsection{Metric Definition}

\paragraph{Normalized target response}
We measure the relative changes in the target \class dimension: $R_{\mathrm{target}}=\Delta \vz_{\mathrm{target}}/\vz_{\mathrm{target}}$, where \(\Delta \vz_{\mathrm{target}}\) is the absolute change after the edited image is decoded and re-encoded, relative to the re-encoded no-op reconstruction, and \(\vz_{\mathrm{target}}\) is the target’s original activation on the input image. We report the round-trip response in the main paper. For signed interventions, the intended response is \(\operatorname{sign}(s)R_{\mathrm{target}}\), where \(s\) is the steering strength, so positive values indicate movement in the requested direction. 

\paragraph{Target selectivity}
We measure the relative changes in the target dimension versus other dimensions. Specifically, we divide the absolute target change by the mean absolute change of the other \class SAE dimensions. Values above one indicate that the target changes more than an average off-target dimension. Larger values indicate better selectivity.

\paragraph{Directional correctness}
We apply both amplified and suppressed interventions on the SAE activations and measure whether the resulting change in the target dimension matches the intended direction. We report the fraction of nonzero-strength interventions for which \(s\Delta \vz_{\mathrm{target}}>0\).

\paragraph{Monotonicity}
Beyond directional correctness, we measure stricter monotonicity, where increasing the signed intervention strength consistently increases the target SAE response. Specifically, we ask that every adjacent response change increases by more than $10^{-6}$. We calculate the metric on the round-trip responses for both patch and \class SAEs and report the fraction of target trajectories with different steering strengths whose response is nondecreasing. 

\paragraph{Strength correlation}
We report the mean Spearman rank correlation between the signed intervention strength and the resulting target response.
For each trajectory from one SAE dimension and one image, we collect the requested strengths \(s\) and the corresponding round-trip target response \(\Delta \vz(s)\). We then compute
\[
\rho_{\text{trajectory}}
=
\operatorname{SpearmanCorr}\bigl(
[s_1,\ldots,s_n],
[\Delta \vz(s_1),\ldots,\Delta \vz(s_n)]
\bigr).
\]Thus, negative strengths should produce negative target responses and positive strengths positive responses. A value near $1$ indicates a monotonic, correctly oriented response curve; $0$ indicates weak or non-monotonic dependence; and negative values indicate an inverted relationship. The correlation is computed on ranks, so it measures ordering rather than assuming a linear response and is therefore less strict than monotonicity.

\section{Paired Benchmark Details}\label{sec:app-pair-benchmark}

This section describes how we build the paired benchmark to evaluate the intervention on single patch-SAE dimensions.

\subsection{Benchmark Constraints}\label{sec:app-benchmark-constraints}
\begin{table}[h]
    \centering
    \small
    \caption{The thresholds for the paired benchmark construction.}
    \label{tab:app-dimension-selection}
    \begin{tabular}{@{}lCC@{}}
    \toprule
        \textbf{Constraint} & \textbf{DINOv3} & \textbf{SigLIP 2} \\
        \midrule
        Max low/high activation ratio & 0.15 & 0.15 \\
        Min high-image percentile & 0.85 & 0.85 \\
        Min activation-percentile gap & 0.60 & 0.50 \\
        Min training activation & 1.0 & 0.1 \\
        Dense-similarity baseline percentile & 99th & 99th \\
    \bottomrule
    \end{tabular}
\end{table}

Our overall target for the selected pairs is that they are \textit{similar in their overall representation but differ clearly along the target SAE feature}. 
First, we require each pair to have dense-embedding similarity above the $99th$ percentile of randomly sampled image pairs, which reduces the risk that the pair can be easily distinguished by coarse differences unrelated to the target feature. 
For SAE activation distinction, we require the high-activation image to lie above the 85th activation percentile, ensuring sufficient signals. The low-to-high activation ratio must be below $0.15$, enforcing a strong relative contrast within the pair. Complementarily, the activation-percentile gap must exceed $0.60$ for DINOv3 and $0.50$ for SigLIP 2, ensuring that the two images are also well separated relative to the feature's overall activation distribution. Finally, we require the high image to reach a minimum training activation of $1.0$ for DINOv3 and $0.1$ for SigLIP 2 to exclude dimensions whose positive examples remain weakly expressed. We select $5$ visually matched pairs for each tested SAE dimension.

\subsection{Evaluation Implementation}
After acquiring the SAE trained on the target dataset, we first filter out dense, dead, and almost dead dimensions, as well as those with stronger activation on backgrounds. We then randomly select $1{,}000$ dimensions from the remaining pool that can produce $5$ image pairs meeting the requirements in \S\ref{sec:app-benchmark-constraints}. Together, we evaluate $5{,}000$ image pairs for each dataset--encoder combination. Given a specific type of feature explanation, the benchmark asks GPT-5.6-Terra to predict the image with higher activations on the target feature. Explanations are presented in two formats: images and text descriptions. 

We test four different groups of image explanations: (1) images with top-ranked activations and the corresponding overlay image showing the patches with high activations, denoted as ``Top''; (2) Top examples and the corresponding reconstructions with the target feature amplified and suppressed; (3) Top examples and bottom examples, \textit{i.e.}, images with the lowest activation in the target feature; (4) the combination of top, bottom examples, and the intervention reconstruction on top examples. When used as the input, we adopt the prompts presented in \S\ref{sec:app-image-prompt}.

In \autoref{tab:main-paired}, we only present the text description generated based on top examples. \autoref{tab:app-text-explanation-comparison} evaluates text descriptions generated based on different image inputs. We use the image input prompts and the text description generation instruction in \S\ref{sec:app-text-prompt} to produce descriptions.

Given the explanation and test image pair, we use the prompt in \S\ref{sec:prediction-prompt} to produce the prediction and calculate the accuracy in predicting the image with higher activation.

\subsection{Prompts for Different Image Input}\label{sec:app-image-prompt}

\begin{tcolorbox}[title=Top input instruction]
  The following examples provide a feature explanation. Each example contains an original image and its activation overlay, where redder regions have higher activation. Infer the visual property consistently highlighted by the overlays.
\end{tcolorbox}

\begin{tcolorbox}[title=Top + intervention input instruction]
  The following examples provide a feature explanation. Each example contains the original image, its activation overlay (redder regions have higher activation), a reconstruction with activation reduced, and a reconstruction with activation increased. Infer the visual property indicated by both the overlay and the consistent change from the reduced to the increased reconstruction.
\end{tcolorbox}

\begin{tcolorbox}[title=Top + bottom input instruction]
  The following examples provide a feature explanation. Each reference contains a high-activating top image and a low-activating bottom image. Infer the visual property that consistently distinguishes the top examples from the bottom examples.
\end{tcolorbox}

\begin{tcolorbox}[title=Top + bottom + intervention input instruction, colback=RoyalBlue!5, colframe=RoyalBlue!50!black]
  The following examples provide a feature explanation. Each reference contains a high-activating top image, its activation overlay (redder regions have higher activation), a low-activating bottom image, a reconstruction from a reduced intervention, and a reconstruction from an increased intervention. Infer the visual property that distinguishes the top examples from the bottom examples and is supported by both the overlays and the consistent reconstruction changes.
\end{tcolorbox}

\subsection{Text Description Generation Prompt}\label{sec:app-text-prompt}

\begin{tcolorbox}[title=Text description generation instruction]
  Return exactly one standalone, concise sentence describing the inferred feature. Be as visually specific as the evidence permits, including concrete color, texture, shape, location, body part, or semantic properties when supported. Do not mention the references, examples, heatmaps, activations, or inference process.
\end{tcolorbox}

\subsection{Prediction Prompt}\label{sec:prediction-prompt}

\begin{tcolorbox}[title=Prediction instruction]
  Given this feature explanation, which image has a higher activation for that feature? Return exactly one letter: A or B.
\end{tcolorbox}

\newpage

\section{Ablation Study}\label{sec:app-ablation}

\subsection{Activation Indication Form}\label{sec:app-activation-form}
\begin{table}[h]
    \centering
    \small
    \caption{Comparison between different forms of indicating the activation region in the top examples. This experiment uses $200$ individual SAE features ($1{,}000$ image pairs) extracted from NABirds using DINOv3 as the encoder.}
    \label{tab:app-activation-form}
    \begin{tabular}{@{}lCC@{}}
    \toprule
        \multirow{2}{*}{\textbf{Activation}} & \multicolumn{2}{c}{\textbf{Explanation Method}} \\
        & \text{Top} & \text{Top + Intervention} \\
        \midrule
        Contour & 67.0 & 70.5 \\
        Overlay & 70.6 & 72.1 \\
    \bottomrule
    \end{tabular}
\end{table}

In the main experiments, we use an overlay to indicate the patches within an image that have high activation in the target sparse feature, with red color for higher activation and blue color for lower activation. 
We compare this design choice with using colored contours around the high-activating patches and report the results in \autoref{tab:app-activation-form}. For both top and top + intervention explanations, the overlay yields a consistent performance advantage, which may be because the overlay provides clearer differentiation between high- and low-activating regions, especially when there is a larger ratio of patch tokens with high activation.

\subsection{Comparisons Using CLIP and IoU Scores}\label{sec:app-clip-iou}
\begin{table}[h]
    \centering
    \small
\begin{minipage}{0.49\textwidth}
    \setlength{\tabcolsep}{4pt}
    \caption{\textbf{Standard similarity metrics do not distinguish explanation quality}. CLIP and IoU scores remain nearly unchanged across explanation settings, despite adding bottom examples and intervention reconstructions that provide qualitatively different semantic evidence. This experiment uses $200$ dimensions ($1{,}000$ image pairs).}
    \label{tab:app-clip-iou}
    \resizebox{\textwidth}{!}{
    \begin{tabular}{@{}cccCC@{}}
    \toprule
        \multicolumn{3}{c}{\textbf{Explanation approach}} & \multicolumn{2}{c}{\textbf{NABirds-DINOv3}} \\
        Top & Bottom & Intervention & \text{CLIP Score} & \text{IoU Score} \\
    \midrule
        \checkmark & & & 
        0.204 & 0.240 \\
        \checkmark & & \checkmark &  
        0.202 & 0.234 \\
        \checkmark & \checkmark & &
        0.202 & 0.248 \\
        \checkmark & \checkmark & \checkmark & 
        0.202 & 0.232 \\
    \bottomrule
    \end{tabular}
    }
\end{minipage}
\hfill
\begin{minipage}{0.46\textwidth}
    \setlength{\tabcolsep}{4pt}
    \caption{\textbf{Comparison between using text and image as the explanation format}. Directly using example images consistently outperforms the text descriptions derived from the same examples. The gap widens as accuracy increases. This experiment uses $200$ dimensions ($1{,}000$ image pairs) from NABirds--DINOv3.}
    \label{tab:app-text-explanation-comparison}
    \resizebox{\textwidth}{!}{
    \begin{tabular}{@{}cccCCC@{}}
    \toprule
        \multicolumn{3}{c}{\textbf{Explanation approach}} & \multicolumn{3}{c}{\textbf{NABirds-DINOv3}} \\
        Top & Bottom & Intervention & \text{Text} & \text{Image} & \Delta \\
    \midrule
        \checkmark & & & 
        62.9 & \mathbf{63.4} & {\color{Green} \uparrow 0.5} \\
        \checkmark & & \checkmark &  
        64.7 & \mathbf{65.4} & {\color{Green} \uparrow 0.7} \\
        \checkmark & \checkmark & &
        67.3 & \mathbf{69.4} & {\color{Green} \uparrow 2.1} \\
        \checkmark & \checkmark & \checkmark & 
        67.1 & \mathbf{72.9} & {\color{Green} \uparrow 5.8} \\
    \bottomrule
    \end{tabular}
    }
\end{minipage}
\end{table}

Prior work employs CLIP and IoU scores to evaluate whether the produced text descriptions can faithfully match the corresponding image regions with high activation~\citep{xu2025deciphering,zhang2025large}. 
CLIP and GroundingDINO models provide accurate recognition and localization for general objects, such as ImageNet categories. However, they may be unreliable when dealing with fine-grained attributes, which are out of their training distribution. We generate the text descriptions given different image inputs using the prompts in \S\ref{sec:app-text-prompt} and report their resulting CLIP and IoU scores in \autoref{tab:app-clip-iou}. The experiment runs on NABirds. Across the four explanation settings, the CLIP score varies only from \(0.202\) to \(0.204\), while IoU ranges from \(0.232\) to \(0.248\), with no consistent improvement as additional explanatory evidence is introduced. These results contrast with our main results in \autoref{tab:main-paired}, where different explanation methods yield large performance differences in identifying feature semantics. This suggests that generic text--image similarity and overlap-based metrics primarily capture coarse semantic correspondence and are insufficiently sensitive to the fine-grained distinctions needed to evaluate sparse-feature explanations.

\subsection{Comparison Between Text and Image Explanation}\label{sec:app-text-explanation-comparison}
We further compare text descriptions generated based on different image explanation inputs in \autoref{tab:app-text-explanation-comparison}. 
Intervention reconstruction improves the prediction accuracy when only top examples are used to derive the feature description. However, the description does not benefit further from intervention when bottom examples are also provided. It can be understood that a concise sentence describing the inferred feature has limited information capacity. In comparison, image explanations, when presented directly to the evaluator, can create more comprehensive feature correspondence with the test image pairs. Notably, as more image explanations are provided, the performance gap between using images directly and using text descriptions also grows. The results support that images provide a more flexible and comprehensive reasoning space compared with text descriptions. 

\subsection{Selectivity on More Steering Strengths}\label{sec:app-more-strengths}

\begin{figure}[h]
    \centering
\begin{subfigure}{0.32\textwidth}
    \includegraphics[width=\textwidth]{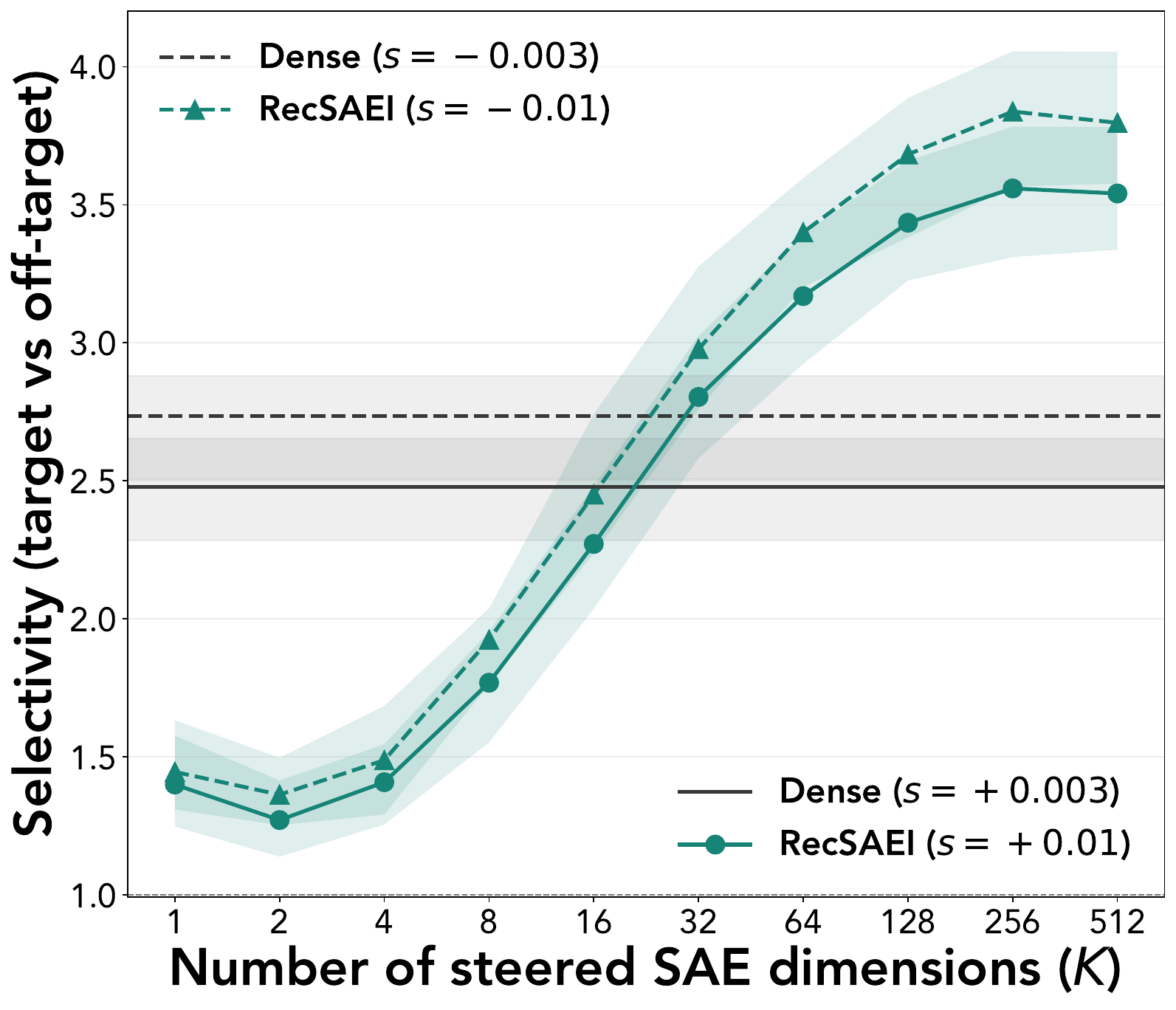}
    \caption{}
    \label{fig:target-response-small}
\end{subfigure}
\hfill
\begin{subfigure}{0.66\textwidth}
    \includegraphics[width=\textwidth]{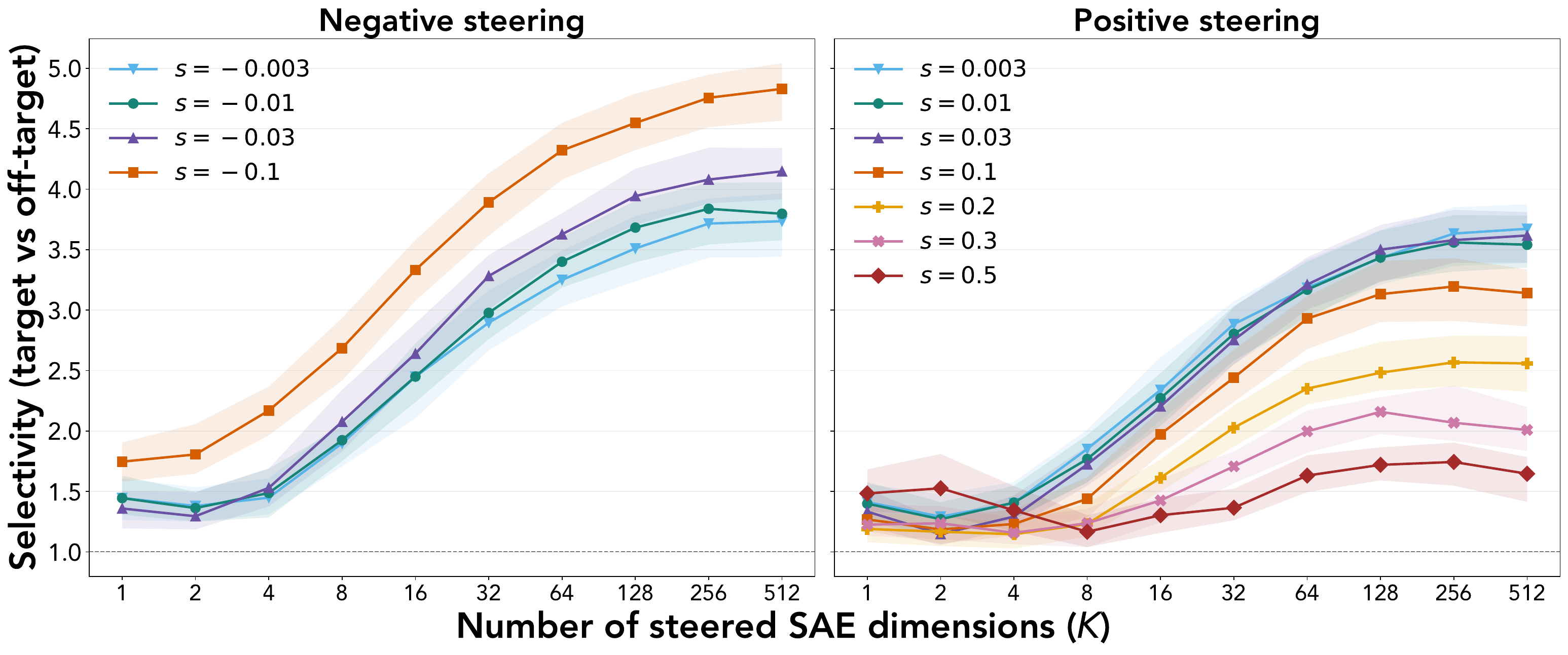}
    \caption{}
    \label{fig:selectivity-strength-varyk}
\end{subfigure}
\caption{\textbf{Effect of steering strength on intervention selectivity}. (a) Selectivity comparison using steering strengths (\(|s|=0.003\) for dense steering and \(|s|=0.01\) for \method) with roughly matched target response. \method remains more selective across different numbers of source dimensions \(K\). (b) Selectivity of \method across intervention strengths. For negative steering, stronger interventions consistently increase selectivity, whereas positive steering achieves its highest selectivity at relatively small strengths and becomes less selective as the intervention grows stronger. Shaded regions indicate variability across evaluated examples.}
\end{figure}

We further study how steering strength affects the selectivity of SAE intervention. 
\autoref{fig:target-response-small} provides an additional comparison with dense steering. In \autoref{fig:selectivity-k}, we compare dense and \method using fixed strengths of \(0.01\) and \(0.03\), respectively. Because the two steering spaces can produce different response magnitudes at the same nominal scale, here we choose another group of strengths that yield approximately matched target responses (\(|s|=0.003\) for dense steering and \(|s|=0.01\) for \method). Under this response-matched setting, \method still achieves higher selectivity as \(K\) increases over $32$, indicating a consistent advantage of the proposed \method.

\autoref{fig:selectivity-strength-varyk} varies the intervention strength \(s\) while increasing the number of attributed patch SAE dimensions \(K\). For negative steering, stronger interventions consistently produce greater separation between the target \class feature and off-target features. This trend remains stable across \(K\), with selectivity generally increasing as more attributed source dimensions are incorporated. 
For positive steering, relatively small strengths, such as \(s=0.003\), \(0.01\), and \(0.03\), produce similar and consistently high selectivity as \(K\) increases. In contrast, further increasing the strength progressively reduces selectivity: although the target feature continues to be promoted, the intervention increasingly affects off-target features as well. Overly strong interventions move the representation beyond the training distribution and introduce broader visual changes. 

\subsection{Reconstruction Using 128 Patch SAE Dimensions}
\begin{figure}[h]
    \centering
    \includegraphics[width=\linewidth]{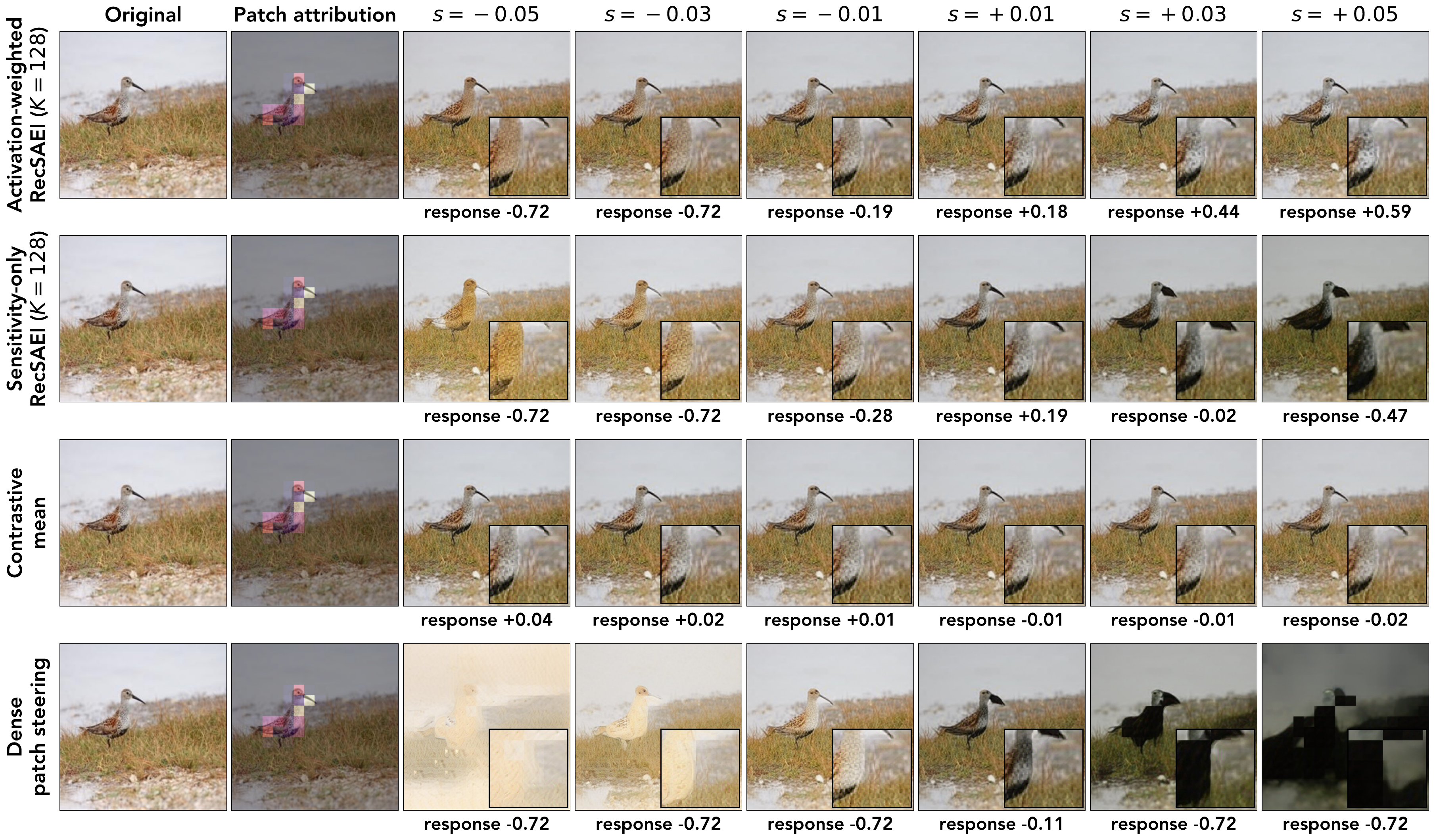}
    \caption{\textbf{Comparison of intervention reconstructions across intervention strengths}. We vary the steering strength \(s\) from negative to positive while keeping the intervention region fixed by the patch attribution. The first row uses our activation-weighted SAE steering with \(K=128\) source features. The remaining rows replace the steering direction with sensitivity-only SAE aggregation, a contrastive-mean direction, and dense patch steering, respectively. The value below each reconstruction reports the target \class feature response.}
    \label{fig:app-selection-ablation-reconstruction}
\end{figure}

We qualitatively compare dense steering and SAE steering using $256$ patch-SAE features in \autoref{fig:reconstruction-comparison}. Here, we also show the comparison using $128$ patch-SAE features in \autoref{fig:app-selection-ablation-reconstruction}. We derive similar conclusions as in the main paper: Dense patch steering is markedly more sensitive to intervention strength than the activation-weighted SAE direction. Even at moderate strengths, the visual changes extend beyond the attributed appearance, while stronger interventions rapidly introduce severe color shifts and reconstruction artifacts. At the same time, the target response decreases quickly due to collapsed visual reconstruction. In contrast, \method produces a substantially smoother progression across \(s\), with visual changes remaining concentrated on the relevant appearance and the target response varying more consistently.

\subsection{Attribution Faithfulness}
\begin{table}[h]
\centering
\small
\captionsetup{type=table,hypcap=false}
    \captionof{table}{\textbf{Target \class response after steering different patch SAE dimensions}. Steering top-attributed dimensions produces substantially larger \class activation changes than steering random active or bottom-attributed dimensions.}\label{tab:attribution-faithfulness}
    \begin{tabular}{@{}l|CCC@{}}
    \toprule
        \textbf{Source} & \textbf{Mean absolute} & \textbf{Mean token} & \textbf{Change} \\
        \textbf{dimensions} & \textbf{\class change} & \textbf{intervention} & \textbf{ratio} \\
    \midrule
        Top attributed    & 7.70 & 396.4 & 1.94\% \\
        Random active     & 0.23 & 92.8 & 0.20\% \\
        Bottom attributed & 0.01 & 35.3 & 0.02\%\\
    \bottomrule
    \end{tabular}
\end{table}
To test whether our attribution scores identify patch SAE features that functionally support the target \class feature, we steer dimensions selected by different criteria and measure the resulting target response. As shown in \autoref{tab:attribution-faithfulness}, top-attributed dimensions produce a much larger mean absolute \class change than randomly selected active dimensions and bottom-attributed dimensions. While these dimensions usually have larger activations and contribute larger interventions, their change ratio per absolute intervention is also larger. This large separation indicates that the proposed attribution identifies sparse patch features whose intervention has a substantially stronger effect on the target \class feature. We ablate the attribution approach in \S\ref{sec:app-attribution-ablation}.

\subsection{Comparison between Attribution Methods}\label{sec:app-attribution-ablation}

\begin{figure}[h]
\centering
\begin{subfigure}{0.34\textwidth}
    \centering
    \includegraphics[width=\linewidth]{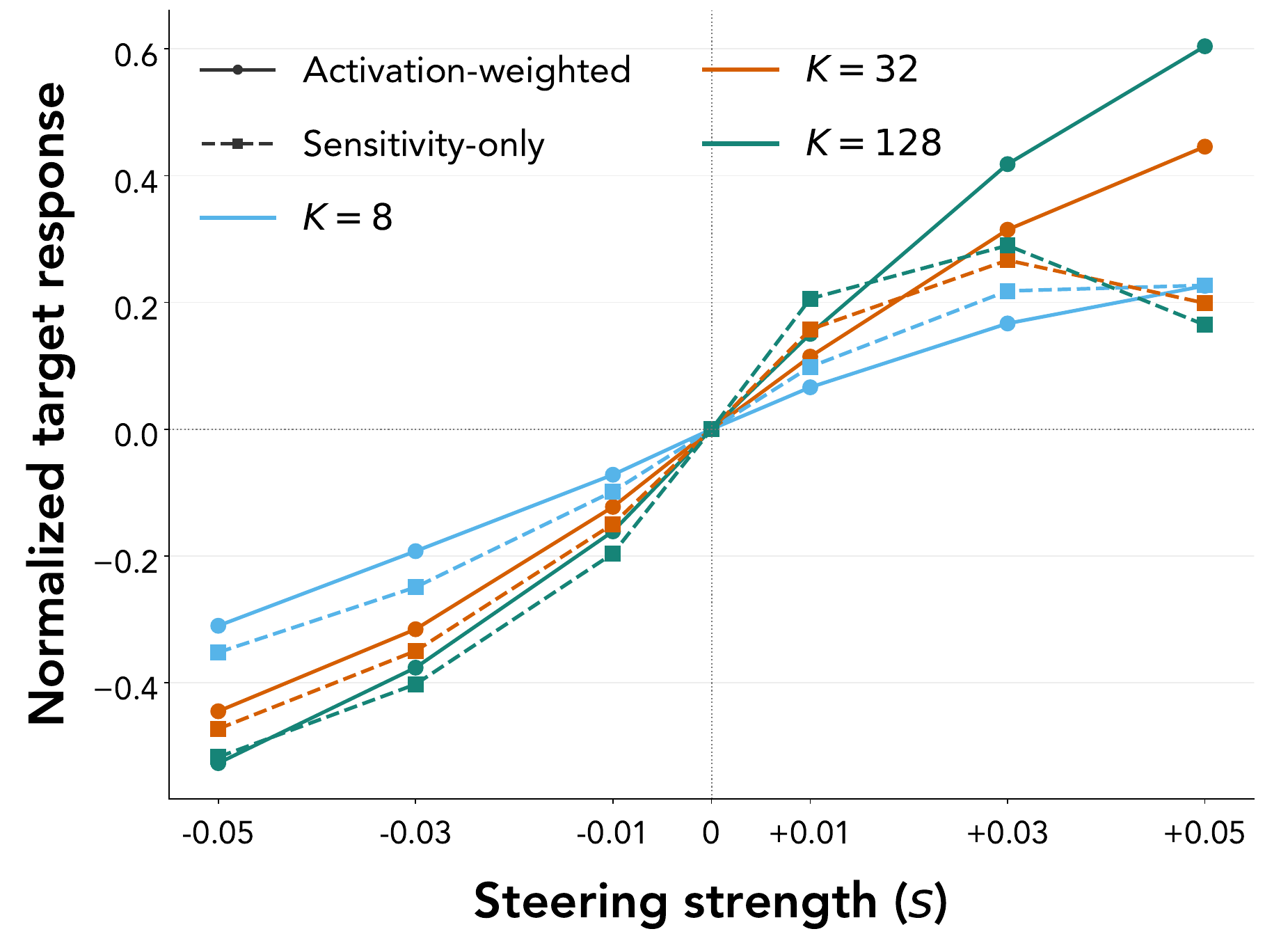}
    \caption{}
    \label{fig:app-attribution-ablation-response}
\end{subfigure}
\begin{subfigure}{0.65\textwidth}
    \centering
    \includegraphics[width=\linewidth]{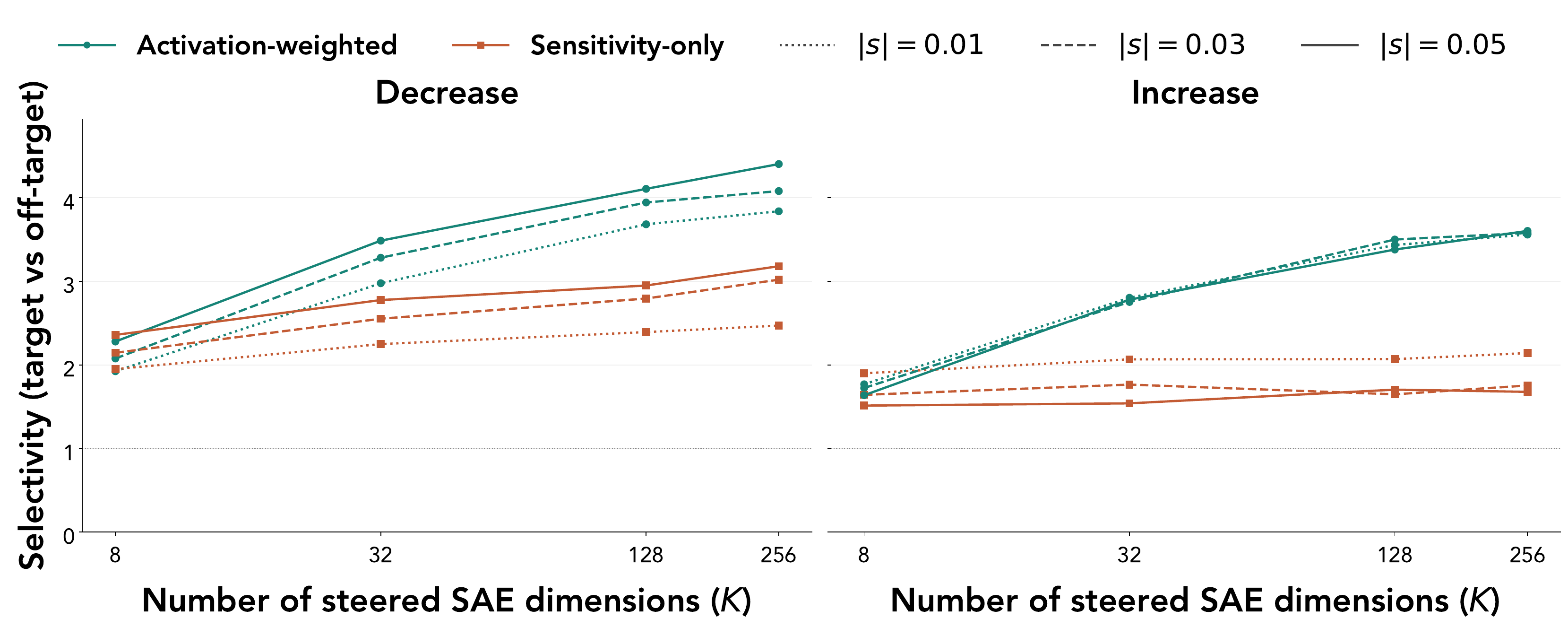}
    \caption{}
    \label{fig:app-attribution-ablation}
\end{subfigure}
\caption{\textbf{Comparison between the proposed activation-weighted dimension attribution and sensitivity-only attribution}.
(a) Normalized target-feature response under different steering strengths. Sensitivity-only attribution yields larger responses at small strengths but also saturates earlier than activation-weighted attribution.
(b) Selectivity for decreasing and increasing interventions. Across steering magnitudes and number of used patch-SAE features, activation-weighted attribution achieves consistently higher selectivity than sensitivity-only steering, with the gap widening as $K$ grows. }
\end{figure}

Our \class intervention constructs its steering direction from patch SAE features using both their sensitivity to the target \class feature and their activation in the current image. Sensitivity captures how strongly changing a patch feature could affect the target, but a highly sensitive feature need not be active in the selected image region. Steering such features may therefore inject directions that are relevant to the target in principle but are not grounded in the image’s current sparse representation. We compare our activation-weighted attribution against a sensitivity-only variant in \autoref{fig:app-attribution-ablation-response} and \autoref{fig:app-attribution-ablation}. 
When only the most sensitive patch-SAE features are used, sensitivity-only attribution produces a slightly larger target response and higher selectivity. However, as the steering strength increases, sensitivity-only attribution saturates more quickly. As shown in the second row of \autoref{fig:app-selection-ablation-reconstruction}, sensitivity-only steering produces a similar visual trend at small intervention strengths but becomes substantially less stable as the magnitude increases. 

The selection diagnostics provide a direct explanation for this behavior. At \(K=128\), \(96.0\%\) of the dimensions selected by sensitivity-only steering are inactive throughout the selected patch mask. Sensitivity-only steering therefore attempts to manipulate features that are largely absent from the sparse representation of the image. As a result, it edits directions that may be locally sensitive in principle but are not grounded in the image’s actual visual evidence, which weakens both controllability and selectivity. In contrast, activation-weighted steering favors source dimensions that are both attributed to the target and active in the selected region, producing a steering direction that is better aligned with the image content and therefore yields stronger, more selective interventions.

\subsection{Comparison with Contrastive Steering}

\begin{figure}[h]
    \centering
\begin{subfigure}{0.64\textwidth}
    \centering
    \includegraphics[width=\linewidth]{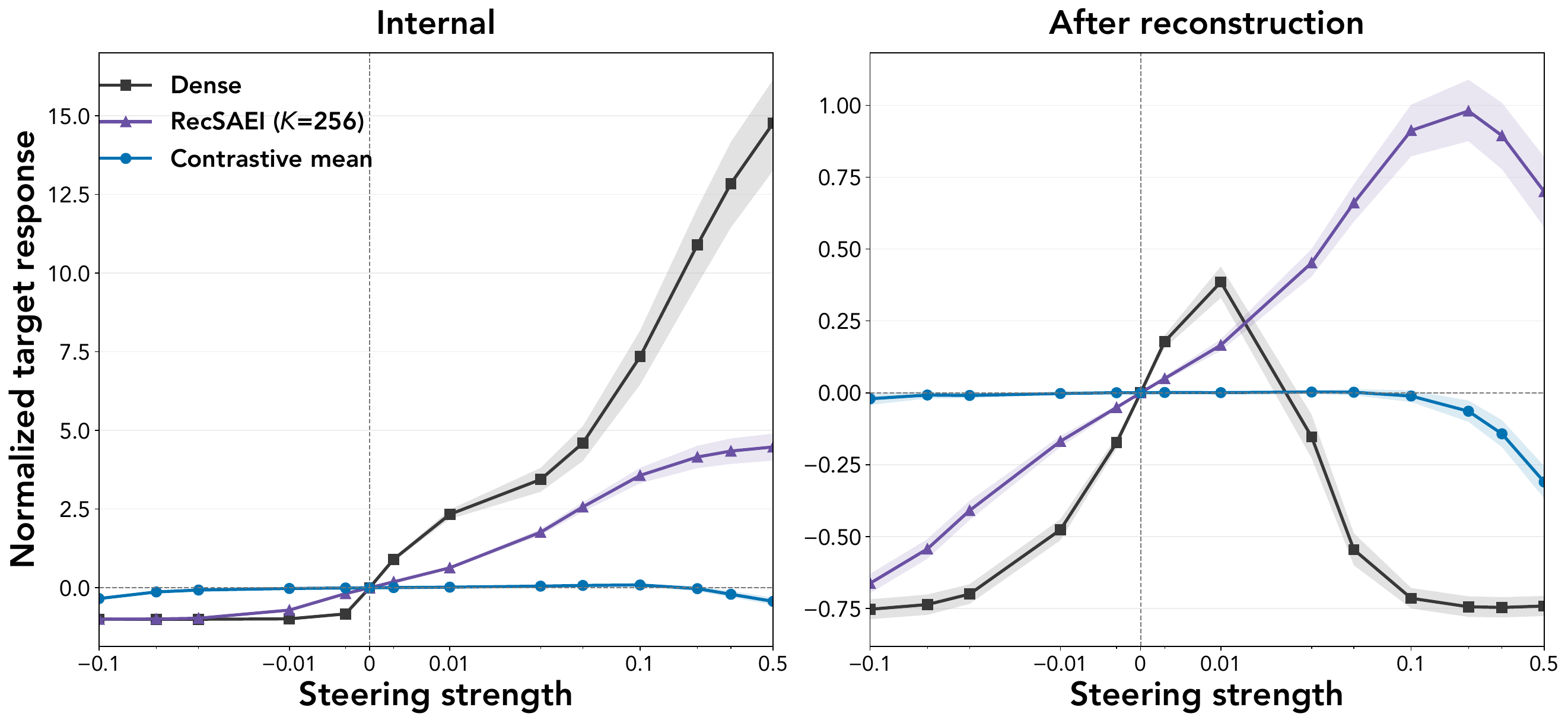}
    \caption{}
    \label{fig:app-internal-vs-reconstructed}
\end{subfigure}
\begin{subfigure}{0.35\textwidth}
    \centering
    \includegraphics[width=\linewidth]{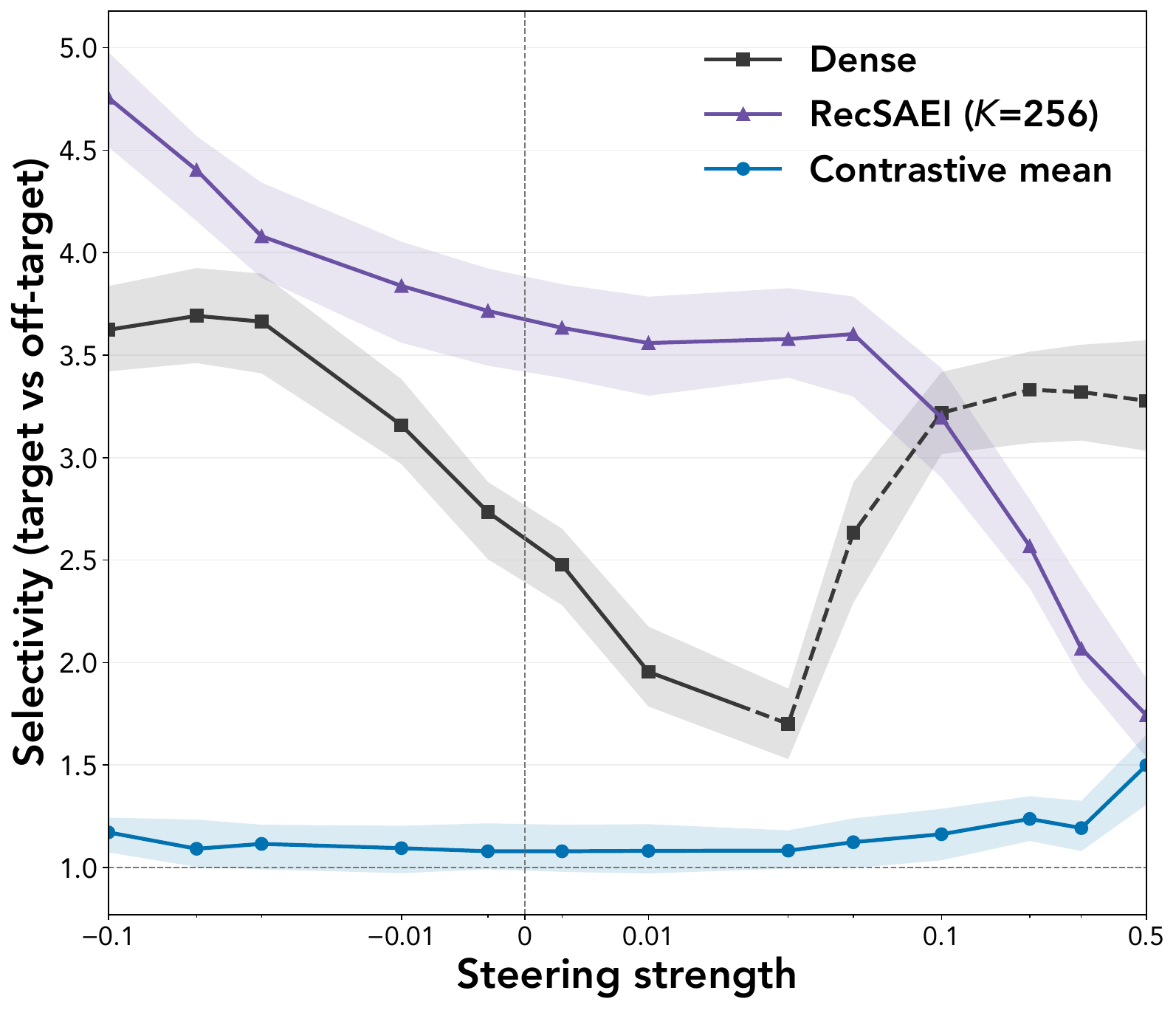}
    \caption{}
    \label{fig:app-selectivity-contrastive}
\end{subfigure}
\caption{\textbf{Comparison between \method and contrastive steering} derived from the mean feature difference between samples with high and low activation.
(a) Normalized target response, measured internally before reconstruction and after reconstruction. Dense steering produces the largest internal response, but much of this effect does not survive reconstruction. In contrast, \method yields a smaller internal response but a substantially stronger and more consistent post-reconstruction target response. The contrastive-mean direction produces almost no controllable response in either space.
(b) Selectivity of the reconstructed interventions. \method is consistently more selective than both dense steering and the contrastive-mean approach.}\label{fig:app-contrastive-comparison}
\end{figure}

Beyond deriving steering directions from gradients, the difference between samples with high and low activation is another option (named ``Contrastive mean''). We compare the intervention effects both before and after reconstruction for dense, SAE, and contrastive mean in \autoref{fig:app-contrastive-comparison}. In the internal representation, dense steering produces the largest target response. However, after reconstruction, the picture changes substantially, leading to decreased target response. By contrast, \method produces a much stronger post-reconstruction response over a wide range of positive strengths. Thus, \method transfers more effectively into the reconstructed image space, and contrastive mean only produces marginal changes in the target response, which is confirmed in \autoref{fig:app-selection-ablation-reconstruction}. Additionally, contrastive-mean stays close to a selectivity of \(1\), indicating that it changes the target little more than off-target features and therefore does not isolate the feature of interest.

\section{More Quantitative Results}\label{sec:app-quantitative}


\subsection{Trace the Steering Changes}
\begin{table}[h]
    \centering
    \small
    \caption{\textbf{Round-trip stability of dense steering and \method under matched injected perturbation magnitude}. We report the resulting patch-feature change after propagation through the remaining encoder block, the target \class-feature change before reconstruction, the target change after reconstruction and re-encoding, and the image-space MSE relative to direct reconstruction. }
    \label{tab:trace-change}
    \resizebox{\textwidth}{!}{
    \begin{tabular}{@{}llCCCCC@{}}
    \toprule
        \textbf{Steering} & \multirow{2}{*}{\textbf{Method}} & \text{\textbf{Injected}} & \text{\textbf{Final}} & \text{\textbf{Internal \class}} & \text{\textbf{Round-trip \class}} & \multirow{2}{*}{\textbf{Image MSE}} \\
        \textbf{Strength} & & \textbf{patch } L_2 & \textbf{patch } L_2 & \text{\textbf{target change}} & \text{\textbf{target change}} & \\
    \midrule
    \multirow{4}{*}{$s=0.01$} & Dense & 140.8 & 4{,}163.9 & 180.9 & 28.7 & 4.6\times 10^{-3} \\
    & \method $K=8$ & 140.8 & 419.3 & 21.1 & 4.7 & 2.3\times 10^{-5} \\
    & \method $K=128$ & 140.8 & 418.7 & 44.8 & 11.4 & 3.3\times 10^{-5} \\
    & \method $K=256$ & 140.8 & 441.5 & 50.7 & 12.6 & 3.9\times 10^{-5} \\
    \midrule
    \multirow{4}{*}{$s=0.03$} & Dense & 422.4 & 16{,}024.6 & 342.5 & -18.5 & 5.5\times 10^{-2} \\
    & \method $K=8$ & 422.4 & 1{,}448.0 & 30.3 & 13.7 & 2.0\times 10^{-4} \\
    & \method $K=128$ & 422.4 & 1{,}456.8 & 91.1 & 31.9 & 2.8\times 10^{-4} \\
    & \method $K=256$ & 422.4 & 1{,}575.9 & 99.8 & 33.9 & 3.3\times 10^{-4} \\
    \bottomrule
    \end{tabular}
    }
\end{table}

Beyond the response in the target \class feature, we further investigate how the applied intervention propagates through the pipeline. As shown in \autoref{tab:trace-change}, we start from the original injected patch embedding intervention. The final patch $L_2$ measures the change in patch embeddings after the final transformer layer. The final patch embeddings also serve as the input to the RAE decoder. For the target \class feature, we measure the change after the final transformer layer (Internal) and the activation after RAE reconstruction and re-encoding using the same encoder. 

Despite receiving the same initial perturbation budget, dense steering is strongly amplified by the remaining encoder computation: its final patch \(L_2\) is roughly an order of magnitude larger than that of \method, and it produces substantially larger internal \class responses. These large internal changes, however, do not translate faithfully through reconstruction. At \(s=0.01\), the dense target response drops from \(180.9\) internally to \(28.7\) after the round trip, while also producing an image MSE over two orders of magnitude larger than \method. The instability becomes more pronounced at \(s=0.03\): although dense steering induces an internal target change of \(342.5\), the round-trip response reverses sign to \(-18.5\), accompanied by a large image deviation. In contrast, \method produces smaller but more stable perturbations: increasing \(K\) strengthens both the internal and round-trip target responses while keeping reconstruction error low. These results suggest that SAE decoder directions produce more stable intervention effects.

\subsection{\class Attribution Statistics}\label{sec:app-cls-attribution-stat}
\begin{table}[h]
    \centering
    \small
    \caption{\textbf{Top attributed patch SAE features} for \class $143$ intervention. We report the top $8$ patch-SAE dimensions ranked by the attribution score used in our activation-weighted steering direction, together with their average activation within the selected patch mask, average absolute sensitivity to the target \class feature, and their rank under sensitivity-only selection. }
    \label{tab:app-cls-attribution-stats}
    \begin{tabular}{@{}LLCCCC@{}}
    \toprule
        \multirow{2}{*}{\textbf{Rank}} & \textbf{Patch-SAE} & \textbf{Attribution} & \textbf{Average} & \textbf{Average absolute} & \textbf{Sensitivity} \\
        & \textbf{dimension} & \textbf{score} & \textbf{activation} & \textbf{sensitivity} & \textbf{Rank} \\
    \midrule
        1 & 14716 & 79.14 & 339.67 & 0.016 & 3   \\
        2 & 10967 & 24.86 & 238.85 & 0.004 & 146 \\
        3 & 9392  & 18.21 & 59.24  & 0.008 & 20  \\
        4 & 7323  & 16.43 & 33.04  & 0.006 & 51  \\
        5 & 4630  & 10.13 & 77.92  & 0.005 & 71  \\
        6 & 12212 & 7.02  & 19.03  & 0.007 & 37  \\
        7 & 5179  & 6.06  & 11.48  & 0.005 & 99  \\
        8 & 15168 & 5.91  & 56.13  & 0.003 & 203 \\
    \bottomrule
    \end{tabular}
\end{table}

The experiments in \S\ref{sec:app-attribution-ablation} compare the proposed activation-weighted attribution with using sensitivity only. We further list the detailed statistics of sensitivity and activation values for the used patch-SAE features in \autoref{tab:app-cls-attribution-stats}.
The sensitivity $q_{p,k}$ measures the alignment between patch-level gradient $\vg_p$ and the decoder direction $\vw_k$. As gradient magnitude is not excluded from the calculation, the sensitivity scale is relatively small. Overall, attribution rank and sensitivity rank differ substantially. Dimension $14716$ is ranked top under both measures. Visualizations indicate that the feature represents the ``elsewhere'' concept~\citep{fel2026into}. The intervention and this attribution analysis further support that the ``elsewhere'' concept is essential in representing the presence of an object.

\section{More Qualitative Results}\label{sec:app-qualitative}

\subsection{\class Dimension Decomposition}\label{sec:app-dimension-decomposition}

\begin{figure}[t]
    \centering
    \includegraphics[width=\linewidth]{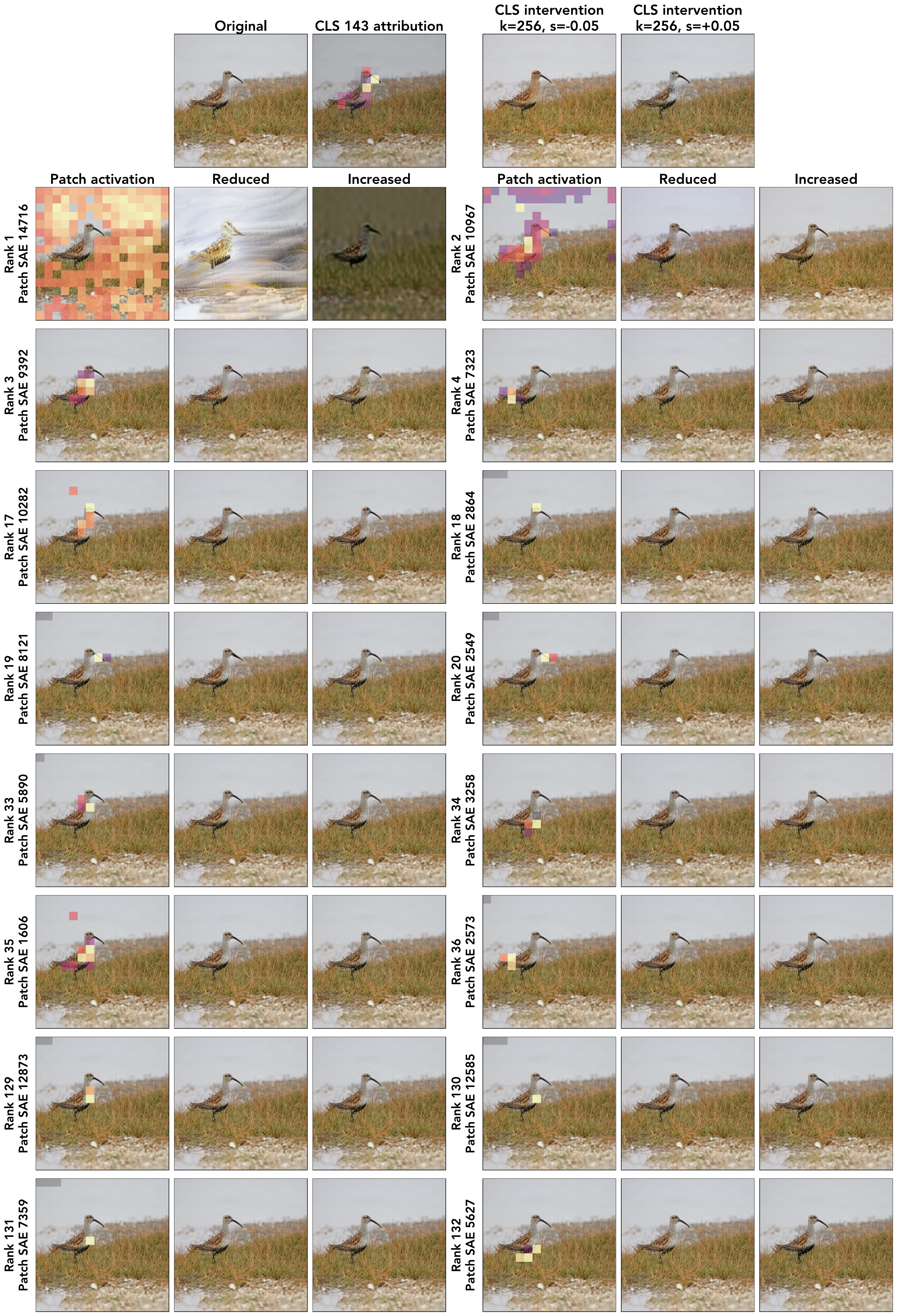}
    \caption{\textbf{Attributed Patch-SAE decomposition of \class SAE feature $143$}. Highly attributed patch-SAE dimensions localize to different image regions and induce distinct visual changes under intervention, revealing the local sparse components that jointly support the global \class feature.}
    \label{fig:app-component-interventions-group}
\end{figure}

\begin{figure}[t]
    \centering
    \includegraphics[width=\linewidth]{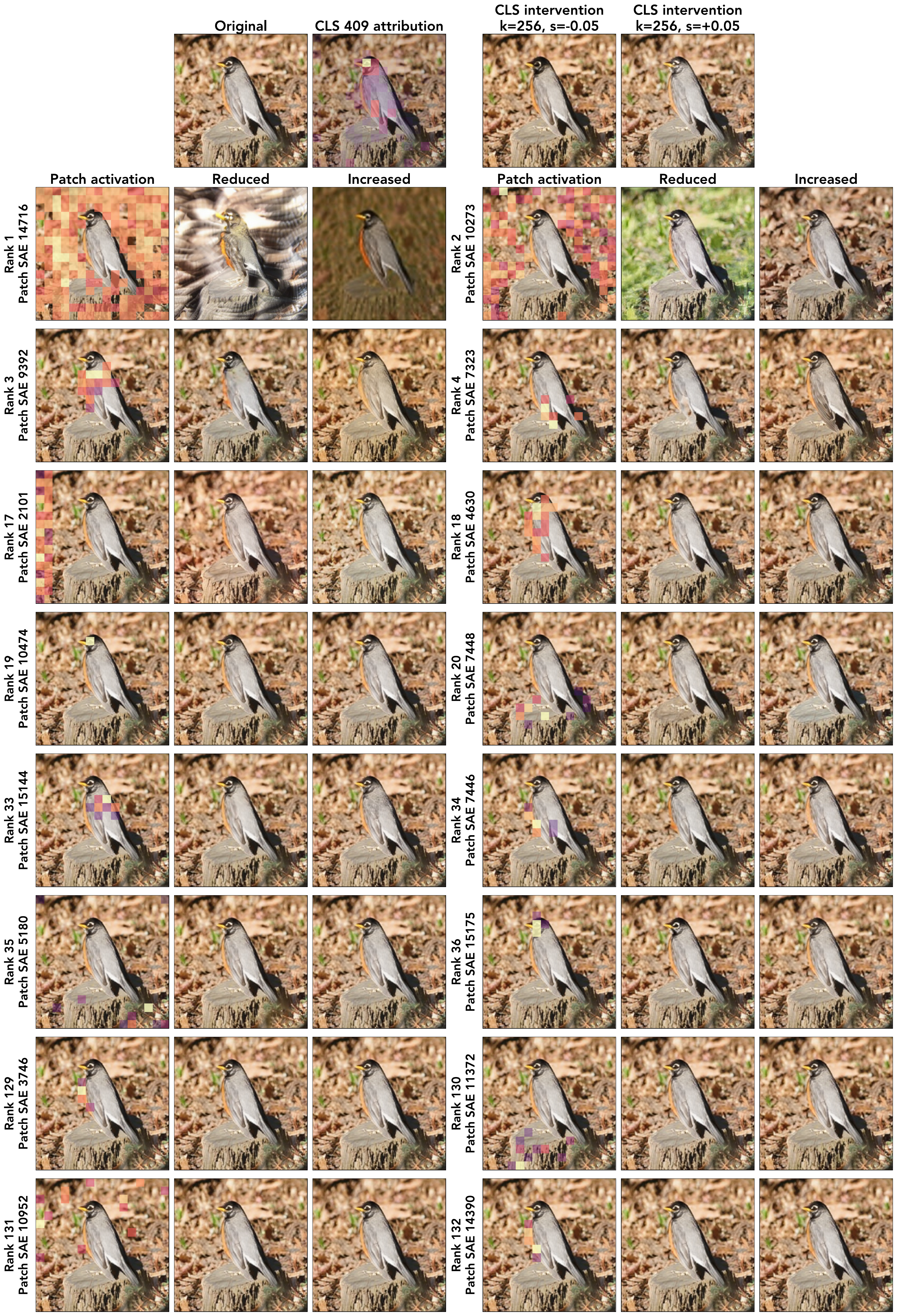}
    \caption{\textbf{Attributed Patch-SAE decomposition of \class SAE feature $409$}. Highly attributed patch SAE dimensions localize to different image regions and induce distinct visual changes under intervention, revealing the local sparse components that jointly support the global \class feature.}
    \label{fig:app-component-interventions-group-1}
\end{figure}

\begin{figure}[t]
    \centering
    \includegraphics[width=\linewidth]{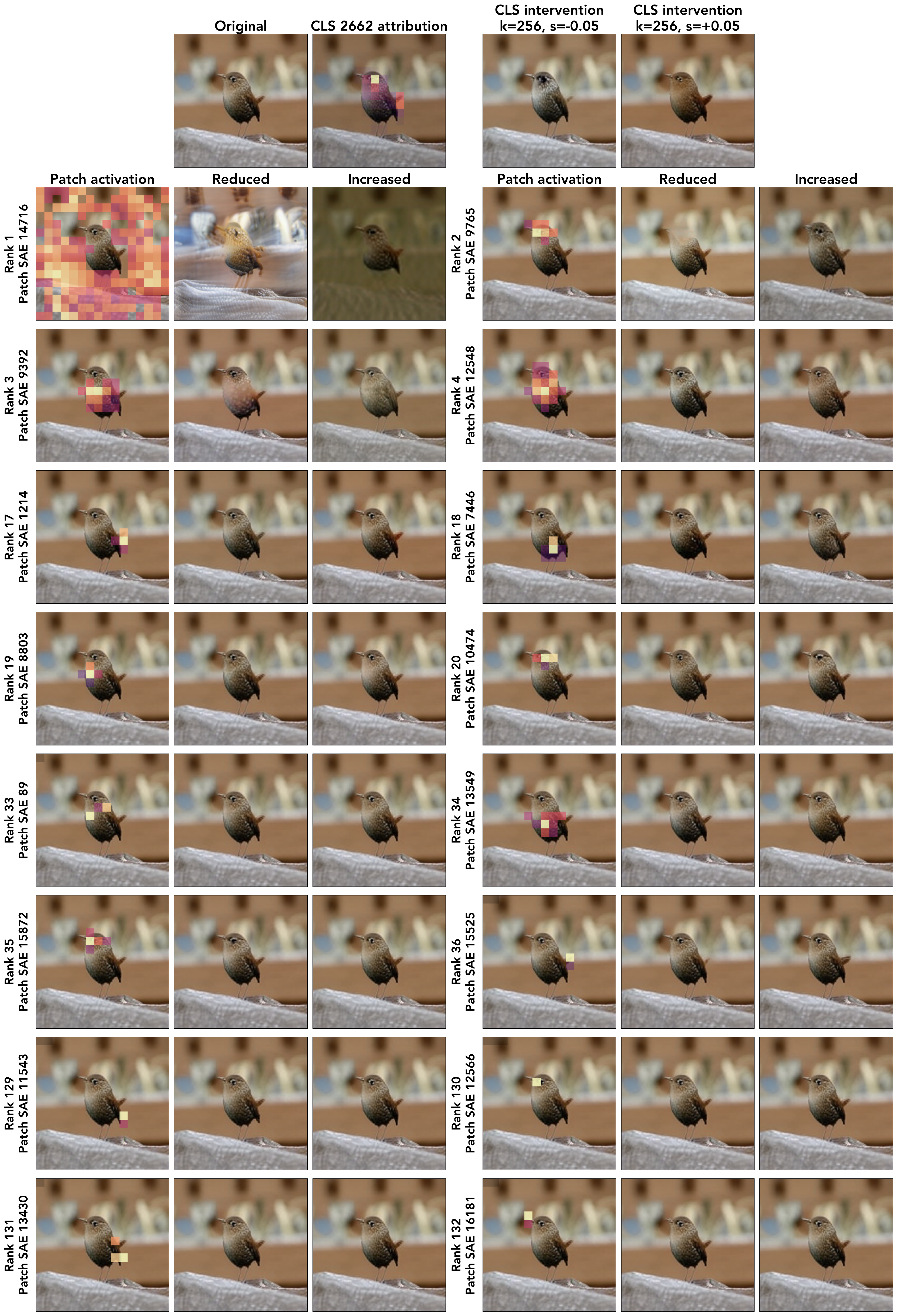}
    \caption{\textbf{Attributed Patch-SAE decomposition of \class SAE feature $2662$}. Highly attributed patch SAE dimensions localize to different image regions and induce distinct visual changes under intervention, revealing the local sparse components that jointly support the global \class feature.}
    \label{fig:app-component-interventions-group-2}
\end{figure}

We present a more comprehensive decomposition result for \class dimension $143$ in \autoref{fig:app-component-interventions-group} and two other \class features in \autoref{fig:app-component-interventions-group-1} and \autoref{fig:app-component-interventions-group-2}.
Across these examples, the highest-ranked patch-SAE feature is consistently dimension $14716$, the ``elsewhere'' concept. The activation overlay shows the normalized activation. Therefore, while the figure suggests the absence of the feature on the object, the activation is merely weaker than background patches. Independently intervening on this feature alters the appearance of nearly the entire image. Activation-weighted attribution identifies the feature as relevant where it overlaps the target \class attribution, but the resulting decoder contribution is applied only within the derived patch mask. We also incorporate more patch-SAE features as $K$ increases, helping attain more stable intervention effects. Nevertheless, the decomposition provides meaningful information on how patch-level features contribute to higher-level \class semantics. 

\subsection{More Examples of Intervention Reconstruction}

\begin{figure}[h]
    \centering
    \includegraphics[width=\linewidth]{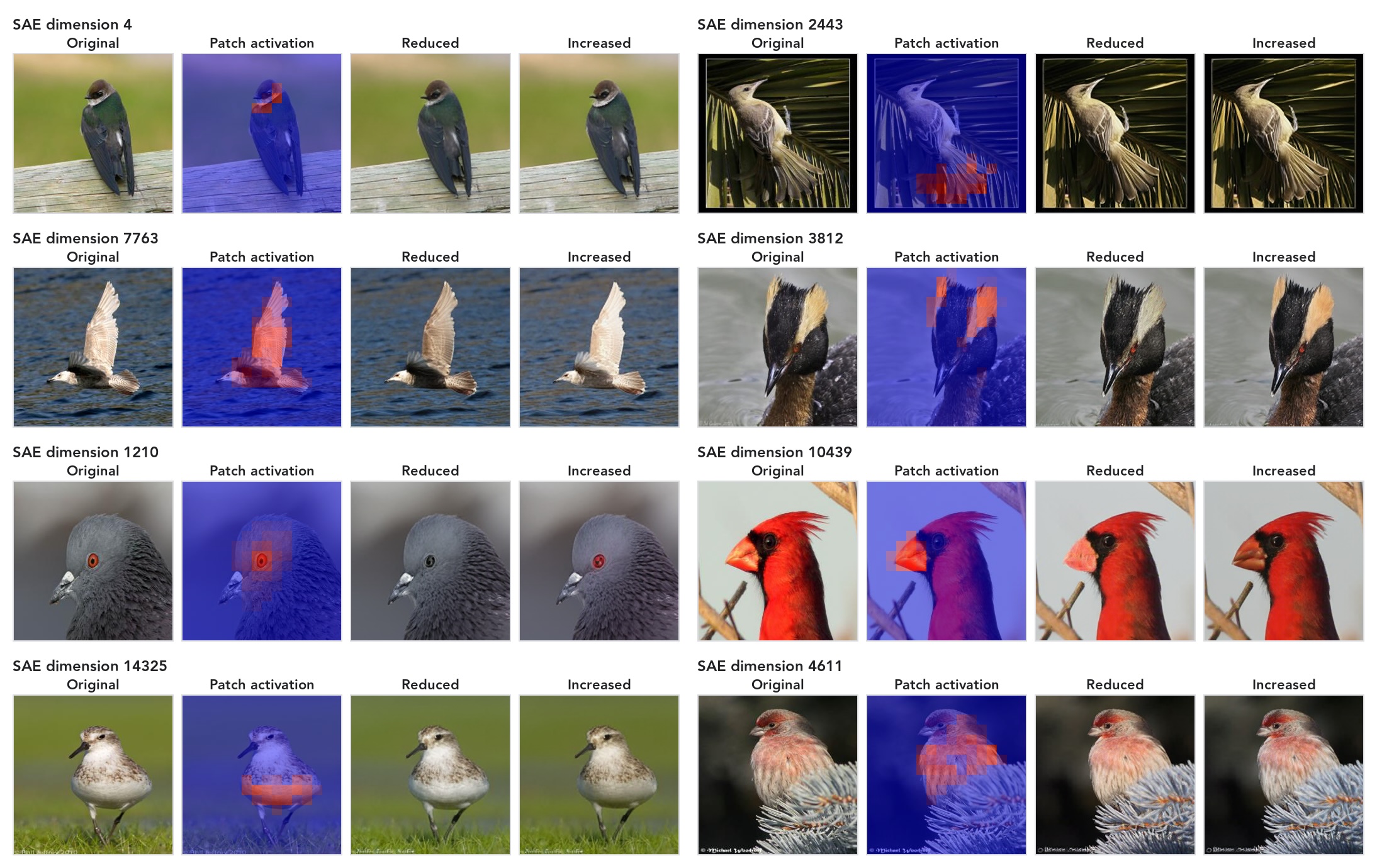}
    \caption{\textbf{Intervention reconstruction examples} on patch-SAE features of NABirds extracted with DINOv3.}
    \label{fig:app-patch-nabirds-dinov3}
\end{figure}

\begin{figure}[h]
    \centering
    \includegraphics[width=\linewidth]{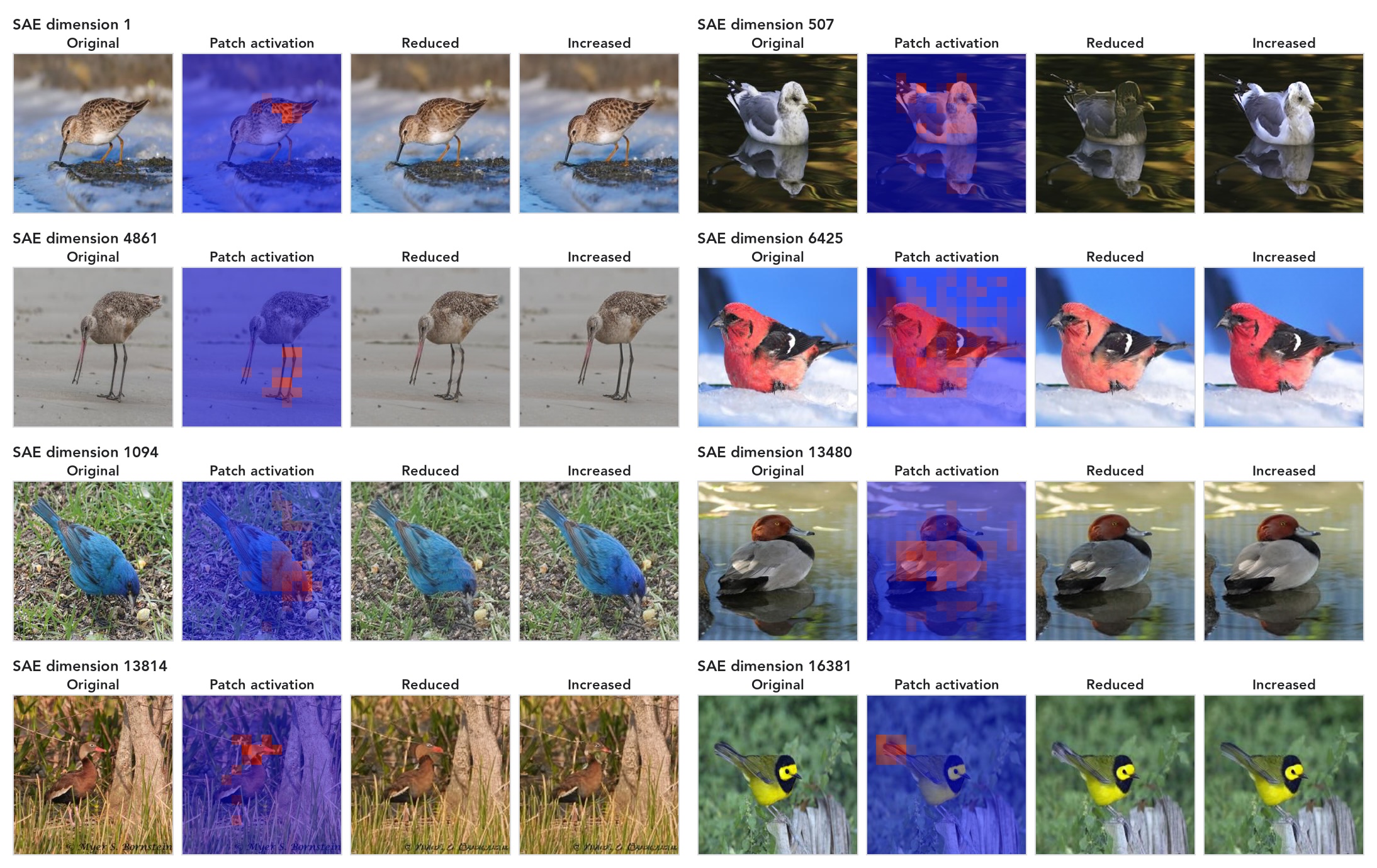}
    \caption{\textbf{Intervention reconstruction examples} on patch-SAE features of NABirds extracted with SigLIP 2.}
    \label{fig:app-patch-nabirds-siglip2}
\end{figure}

\begin{figure}[h]
    \centering
    \includegraphics[width=\linewidth]{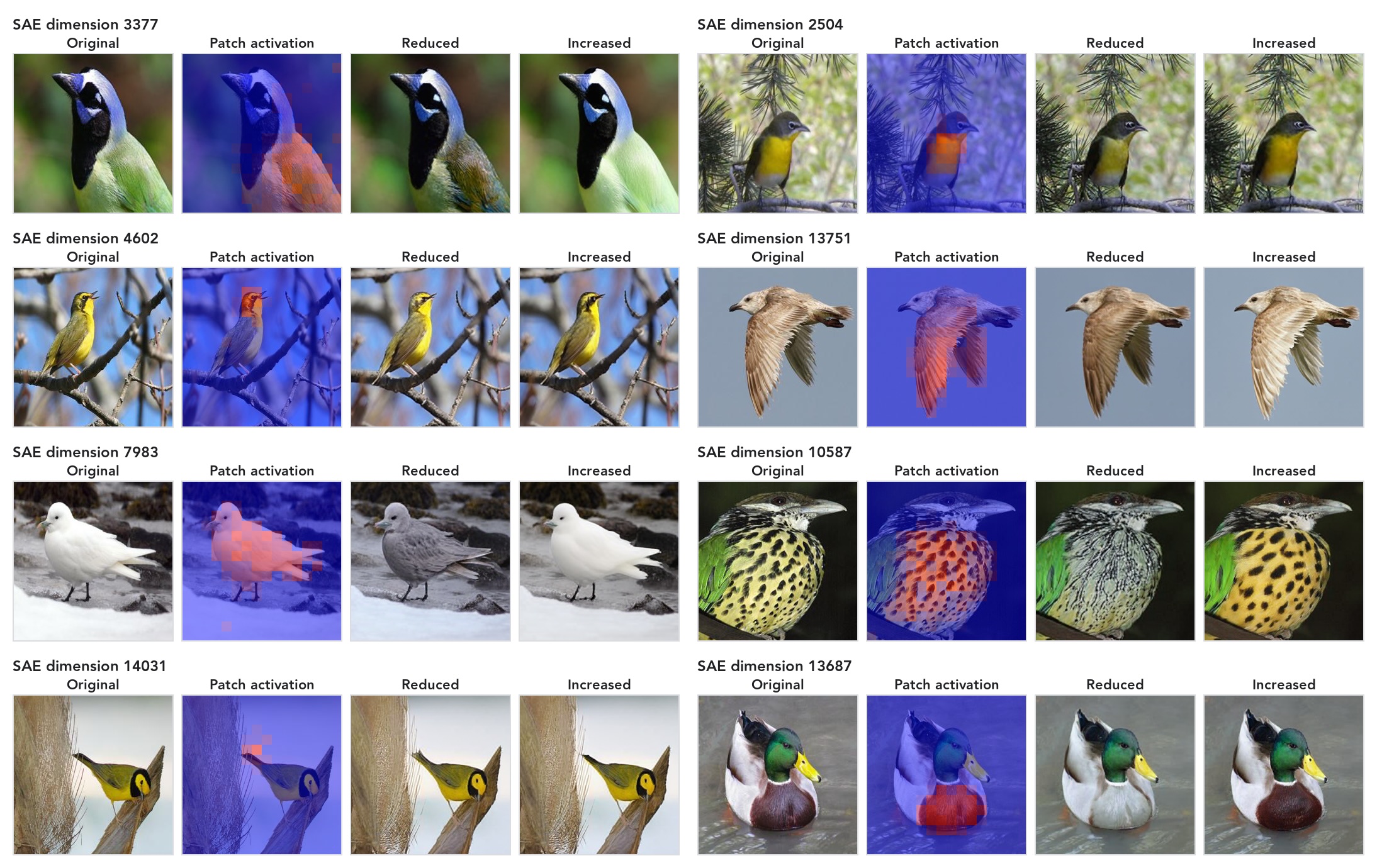}
    \caption{\textbf{Intervention reconstruction examples} on patch-SAE features of CUB extracted with DINOv3.}
    \label{fig:app-patch-cub-dinov3}
\end{figure}

\begin{figure}[h]
    \centering
    \includegraphics[width=\linewidth]{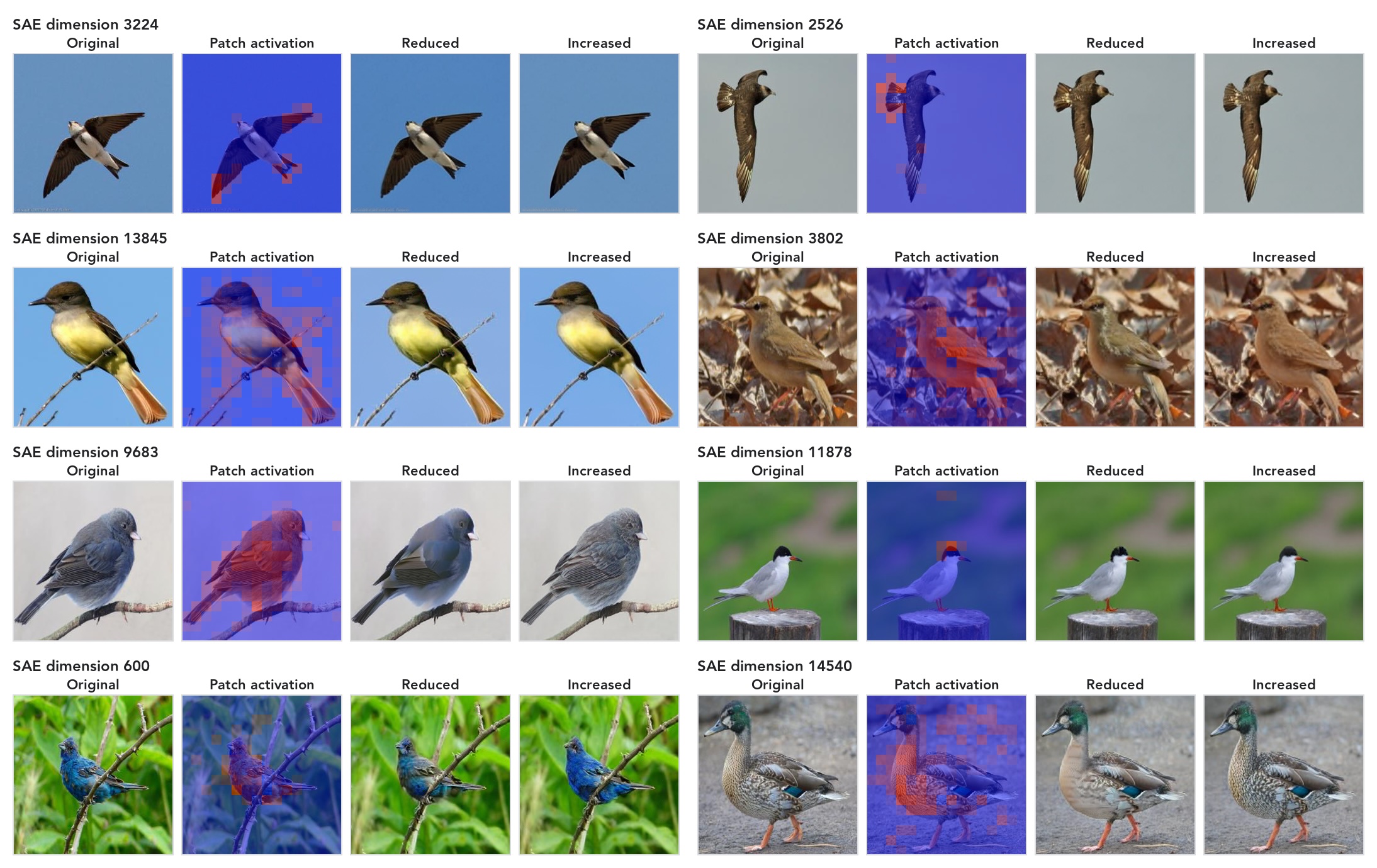}
    \caption{\textbf{Intervention reconstruction examples} on patch-SAE features of CUB extracted with SigLIP 2.}
    \label{fig:app-patch-cub-siglip2}
\end{figure}

\begin{figure}[h]
    \centering
    \includegraphics[width=\linewidth]{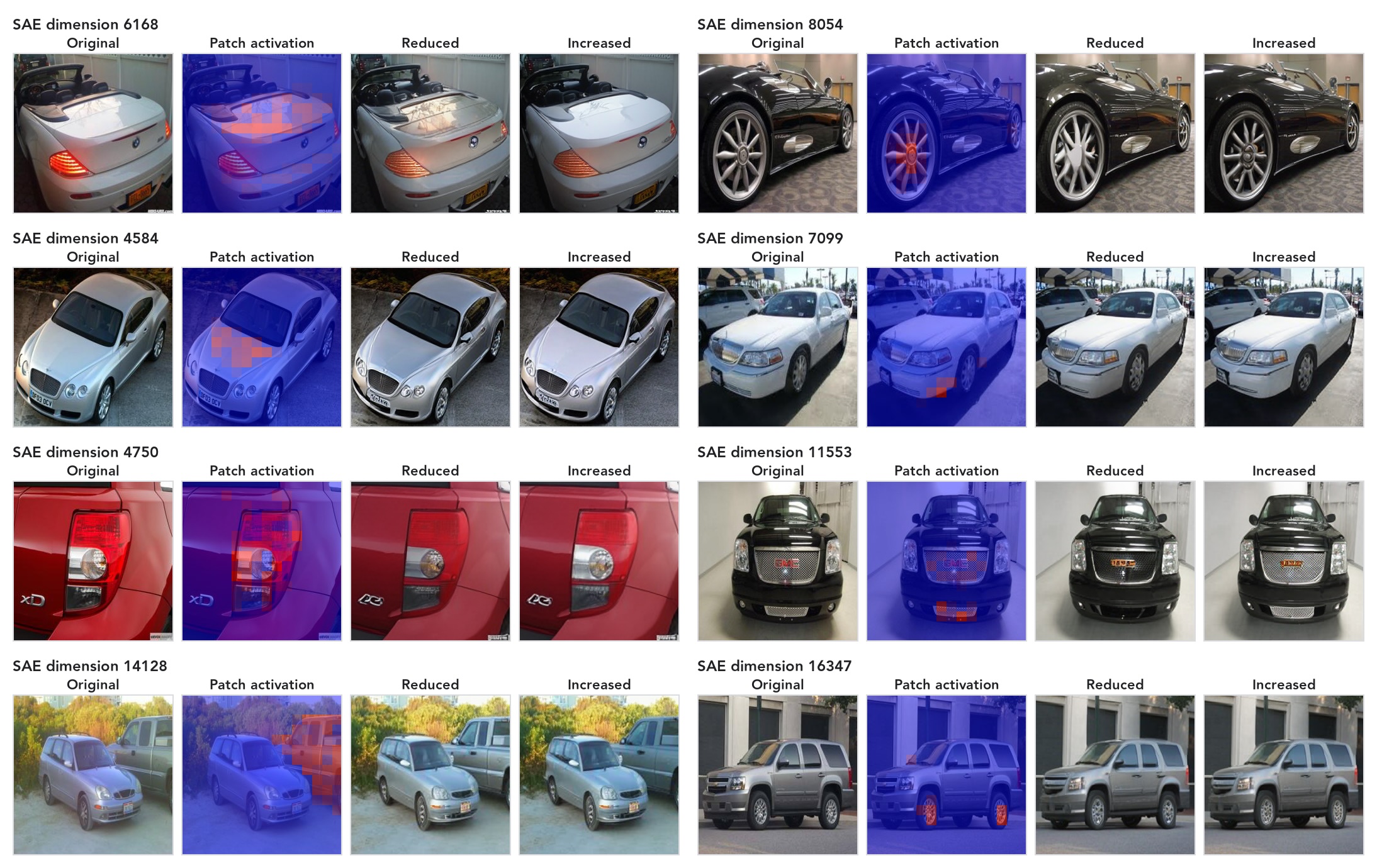}
    \caption{\textbf{Intervention reconstruction examples} on patch-SAE features of Cars extracted with DINOv3.}
    \label{fig:app-patch-cars-dinov3}
\end{figure}

\begin{figure}[h]
    \centering
    \includegraphics[width=\linewidth]{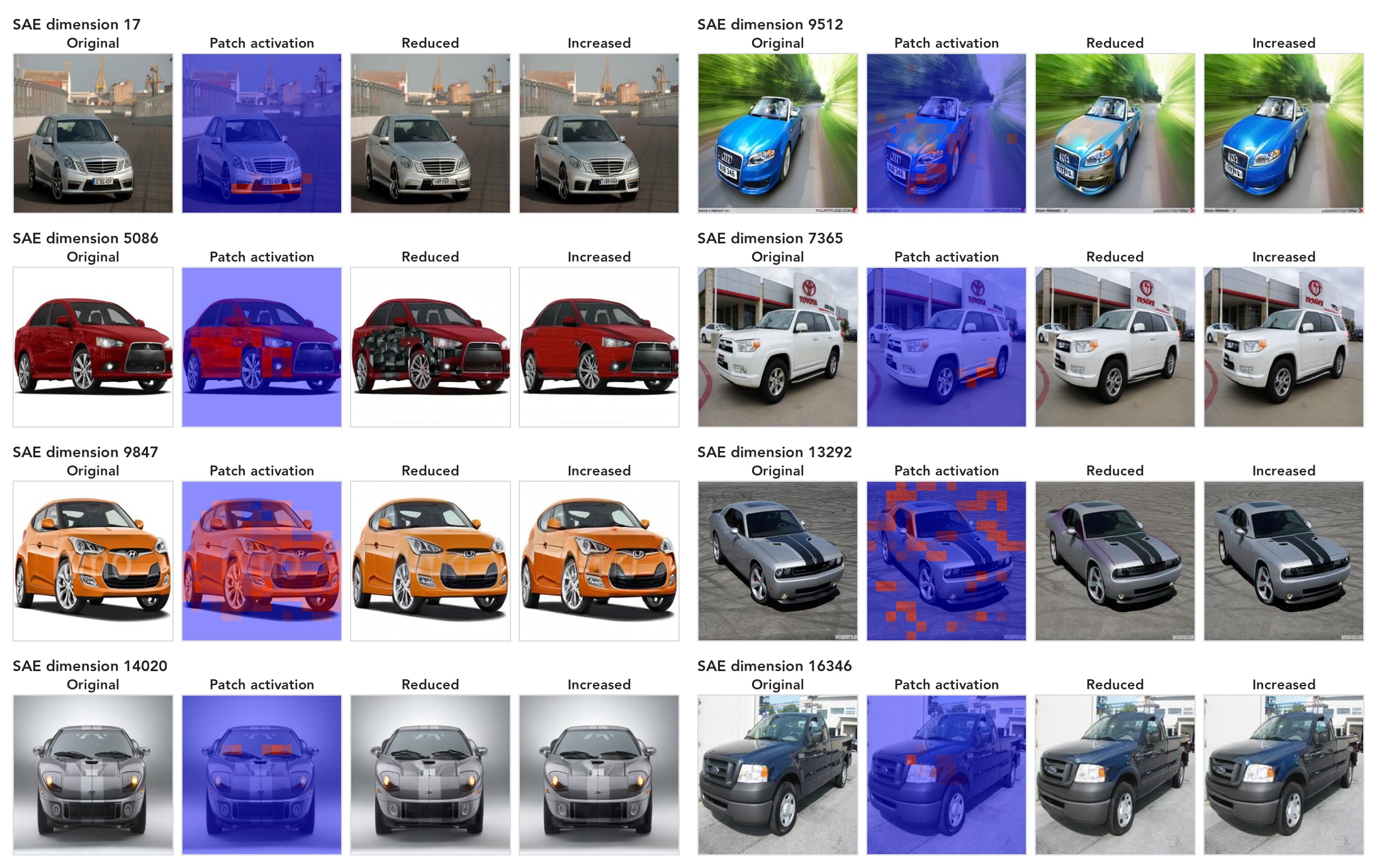}
    \caption{\textbf{Intervention reconstruction examples} on patch-SAE features of Cars extracted with SigLIP 2.}
    \label{fig:app-patch-cars-siglip2}
\end{figure}

\begin{figure}[h]
    \centering
    \includegraphics[width=\linewidth]{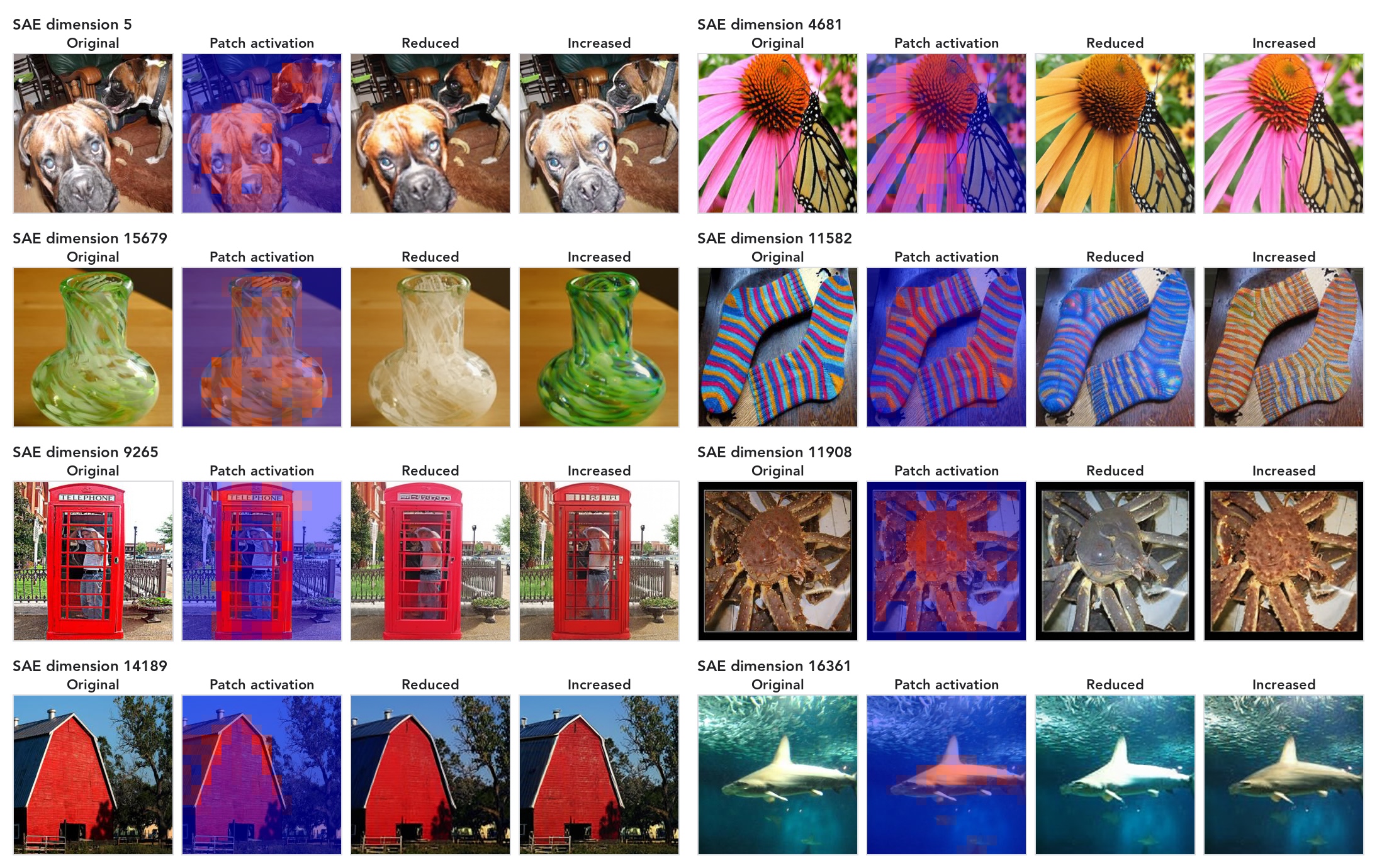}
    \caption{\textbf{Intervention reconstruction examples} on patch-SAE features of ImageNet extracted with DINOv3.}
    \label{fig:app-patch-imagenet-dinov3}
\end{figure}

\begin{figure}[h]
    \centering
    \includegraphics[width=\linewidth]{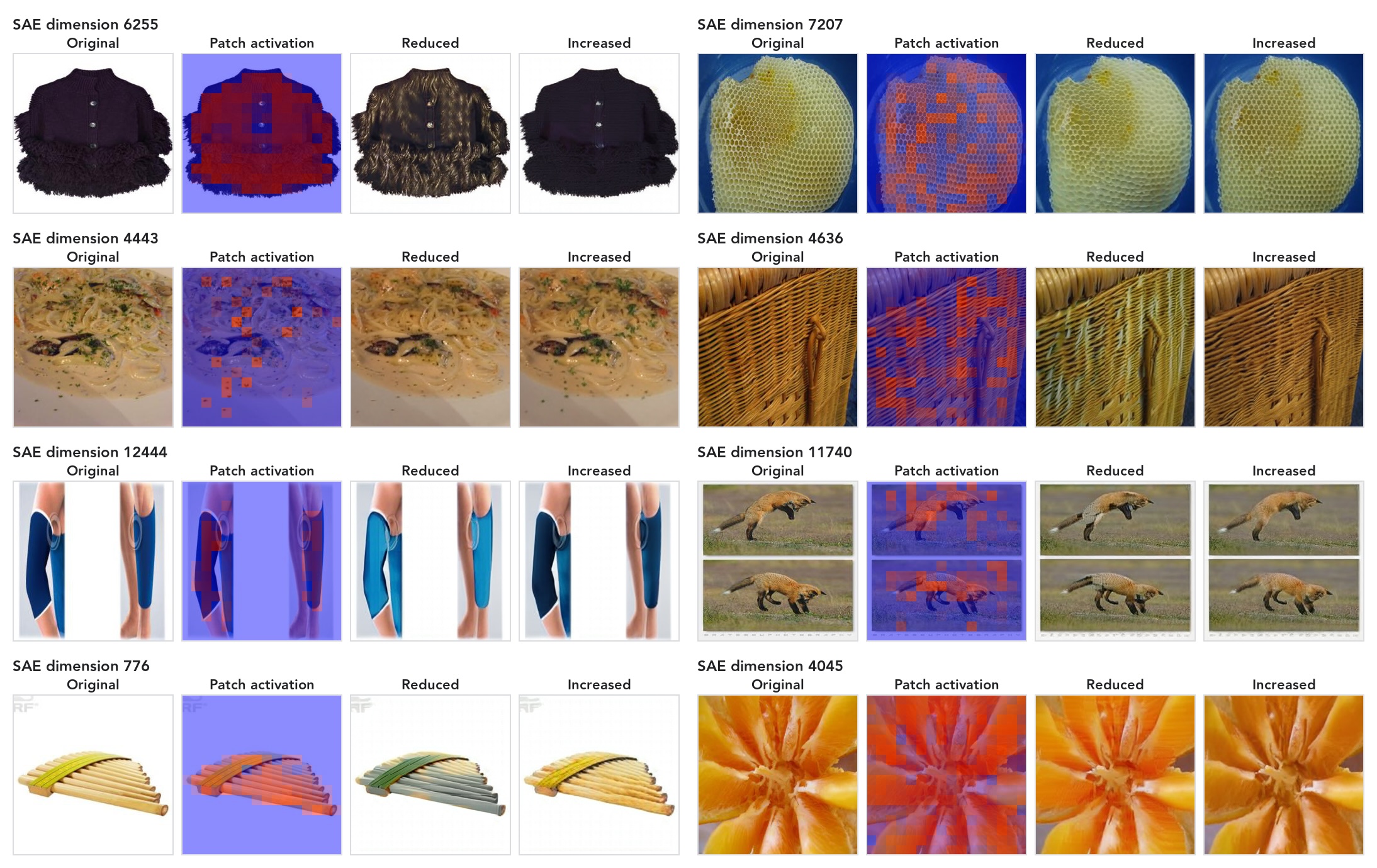}
    \caption{\textbf{Intervention reconstruction examples} on patch-SAE features of ImageNet extracted with SigLIP 2.}
    \label{fig:app-patch-imagenet-siglip2}
\end{figure}

\begin{figure}[h]
    \centering
    \includegraphics[width=0.95\linewidth]{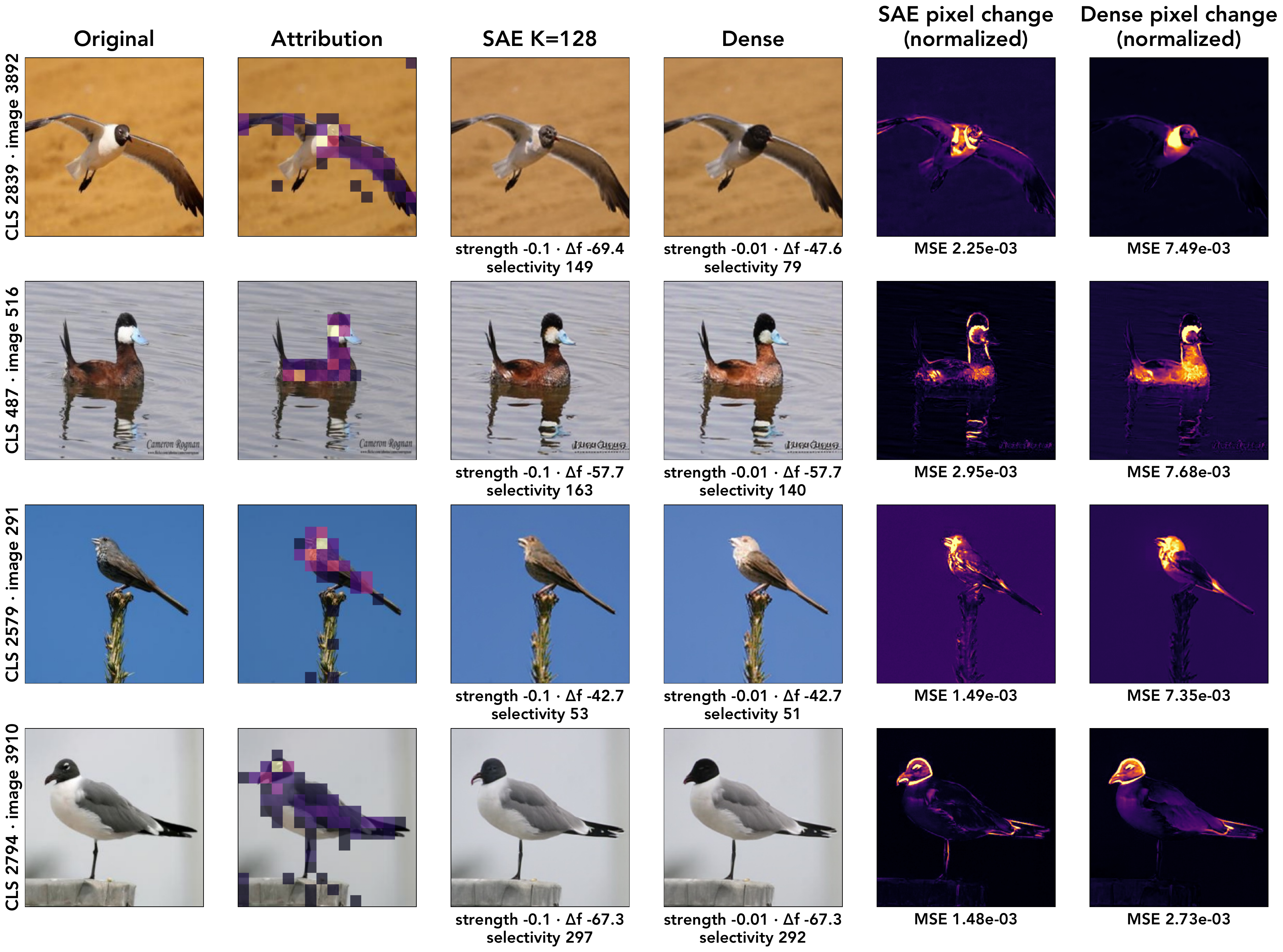}
    \caption{\textbf{Examples of suppressed intervention reconstruction} on \class SAE features of NABirds extracted with DINOv3.}
    \label{fig:app-cls-steering-decrease-example}
\end{figure}

\begin{figure}[h]
    \centering
    \includegraphics[width=0.95\linewidth]{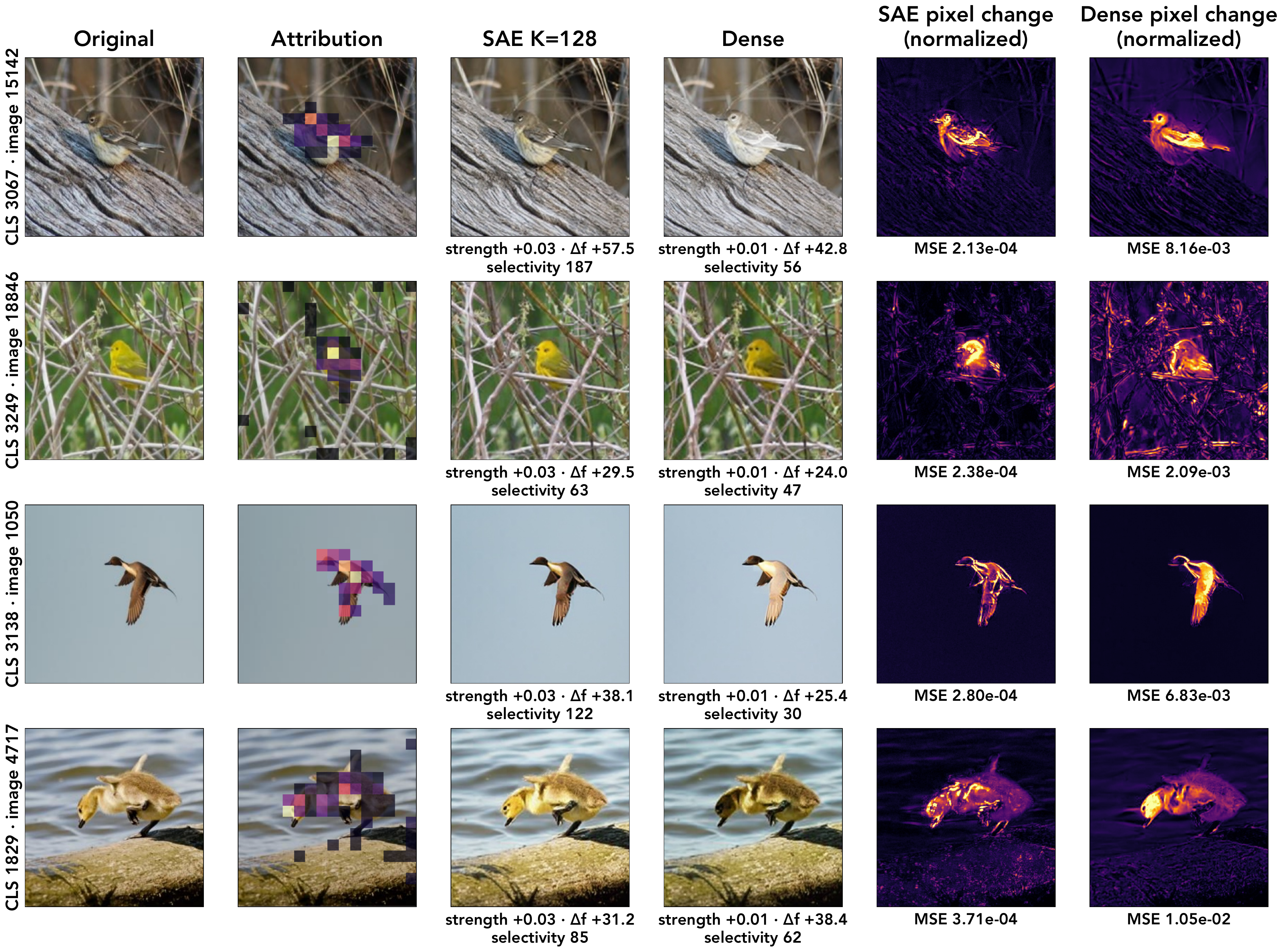}
    \caption{\textbf{Examples of amplified intervention reconstruction} on \class SAE features of NABirds extracted with DINOv3.}
    \label{fig:app-cls-steering-increase-example}
\end{figure}

\begin{figure}[h]
    \centering
    \includegraphics[width=0.95\linewidth]{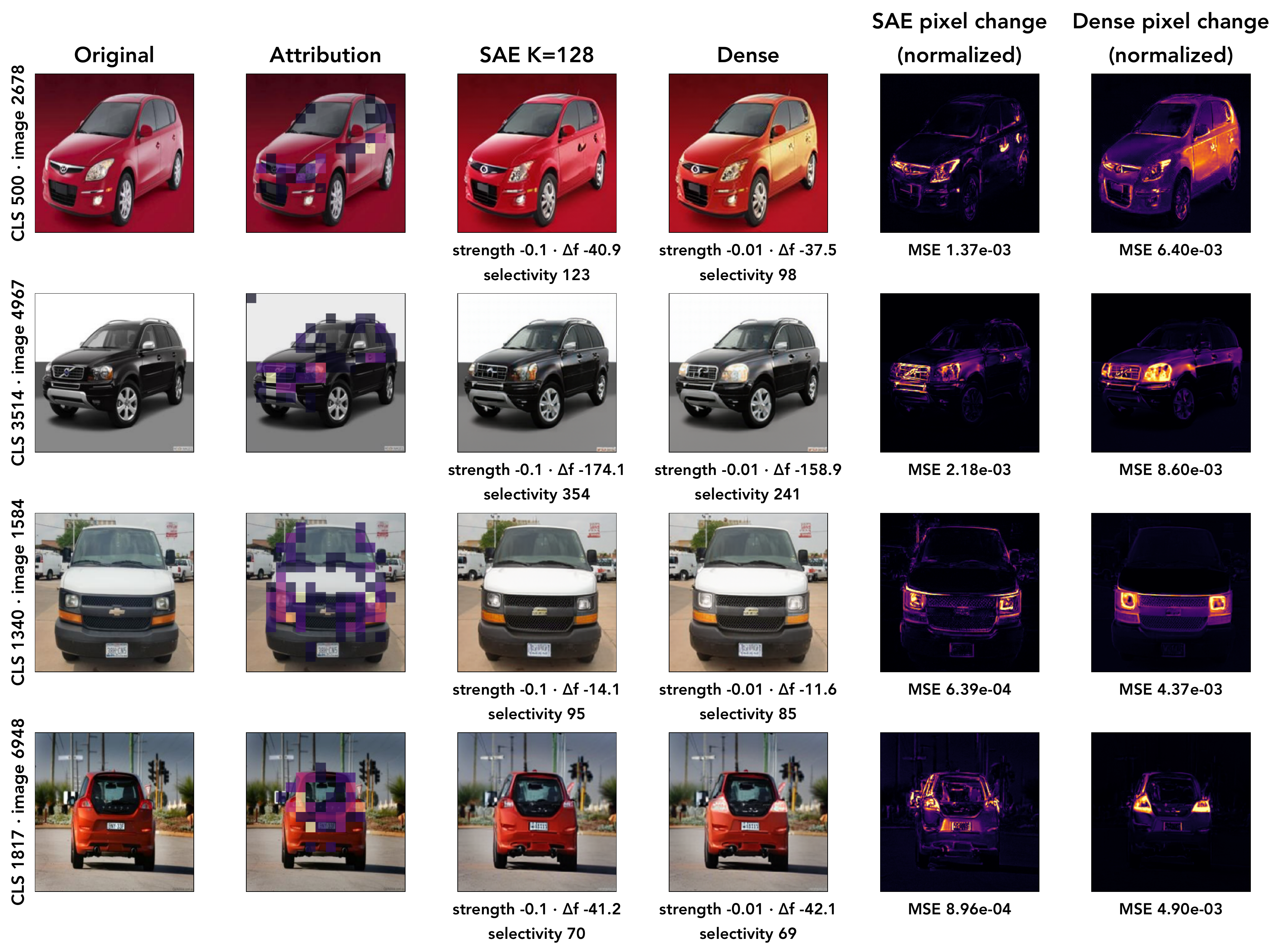}
    \caption{\textbf{Examples of suppressed intervention reconstruction} on \class SAE features of Cars extracted with DINOv3.}
    \label{fig:app-cars-cls-steering-decrease-example}
\end{figure}

\begin{figure}[h]
    \centering
    \includegraphics[width=0.95\linewidth]{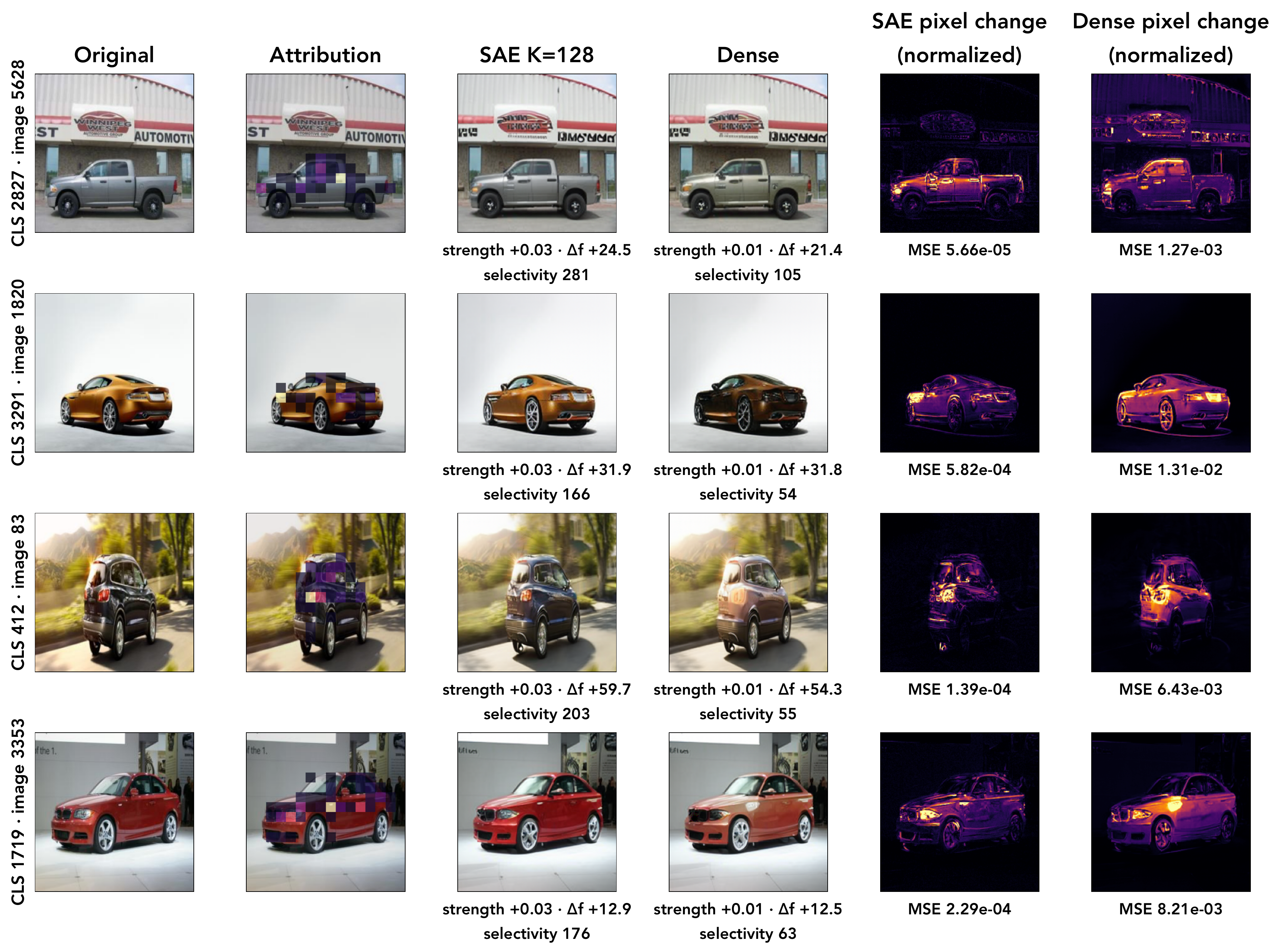}
    \caption{\textbf{Examples of amplified intervention reconstruction} on \class SAE features of Cars extracted with DINOv3.}
    \label{fig:app-cars-cls-steering-increase-example}
\end{figure}

We present more examples of intervention reconstruction on patch-SAE features from \autoref{fig:app-patch-nabirds-dinov3} to \autoref{fig:app-patch-imagenet-siglip2}, and on \class SAE features from \autoref{fig:app-cls-steering-decrease-example} to \autoref{fig:app-cars-cls-steering-increase-example}.

\subsection{More Consistency Examples for \class SAE Intervention}
\begin{figure}[h]
    \centering
\begin{subfigure}{0.49\textwidth}
    \centering
    \includegraphics[width=\textwidth]{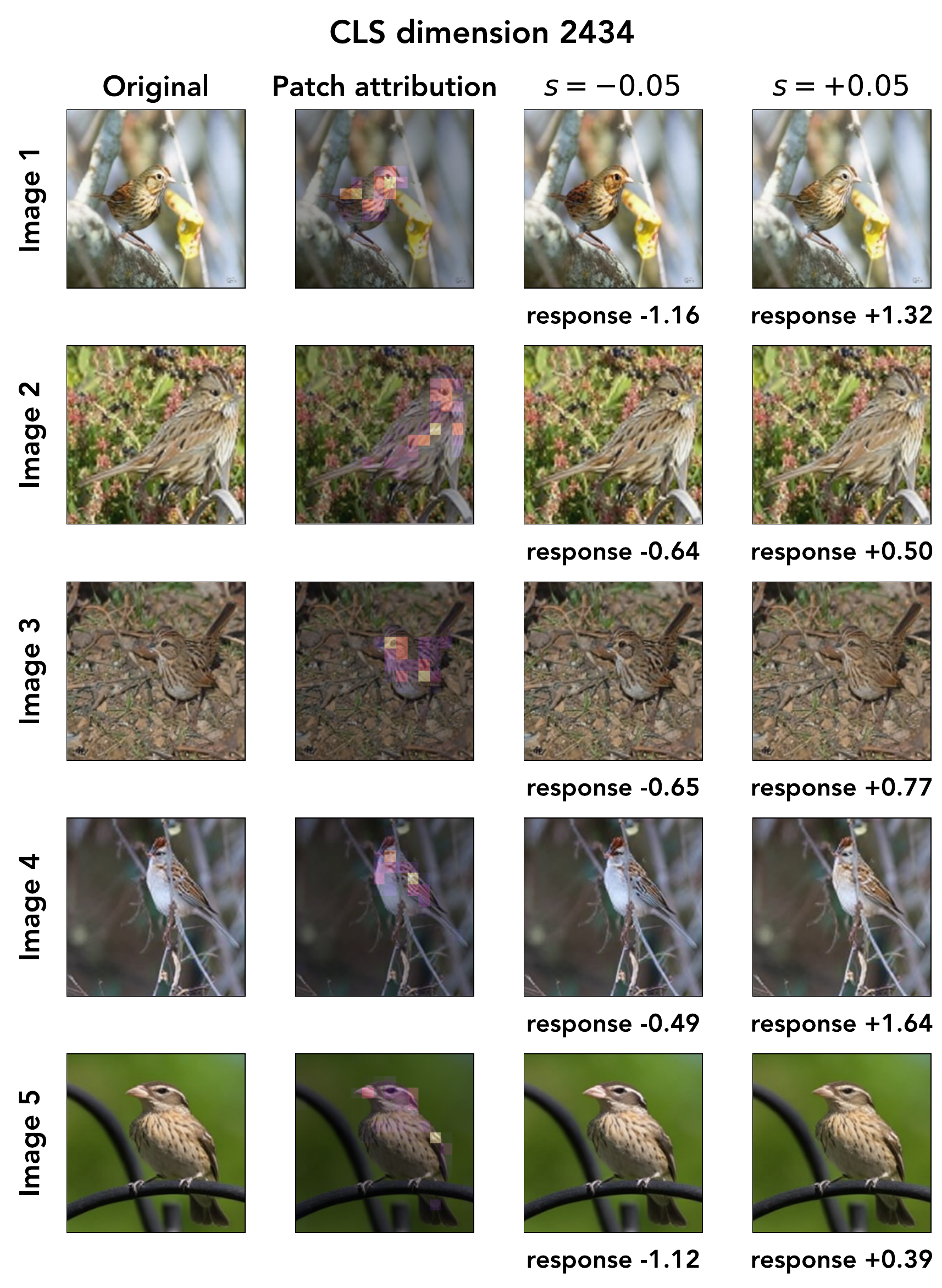}
    \caption{}
\end{subfigure}
\hfill
\begin{subfigure}{0.49\textwidth}
    \centering
    \includegraphics[width=\textwidth]{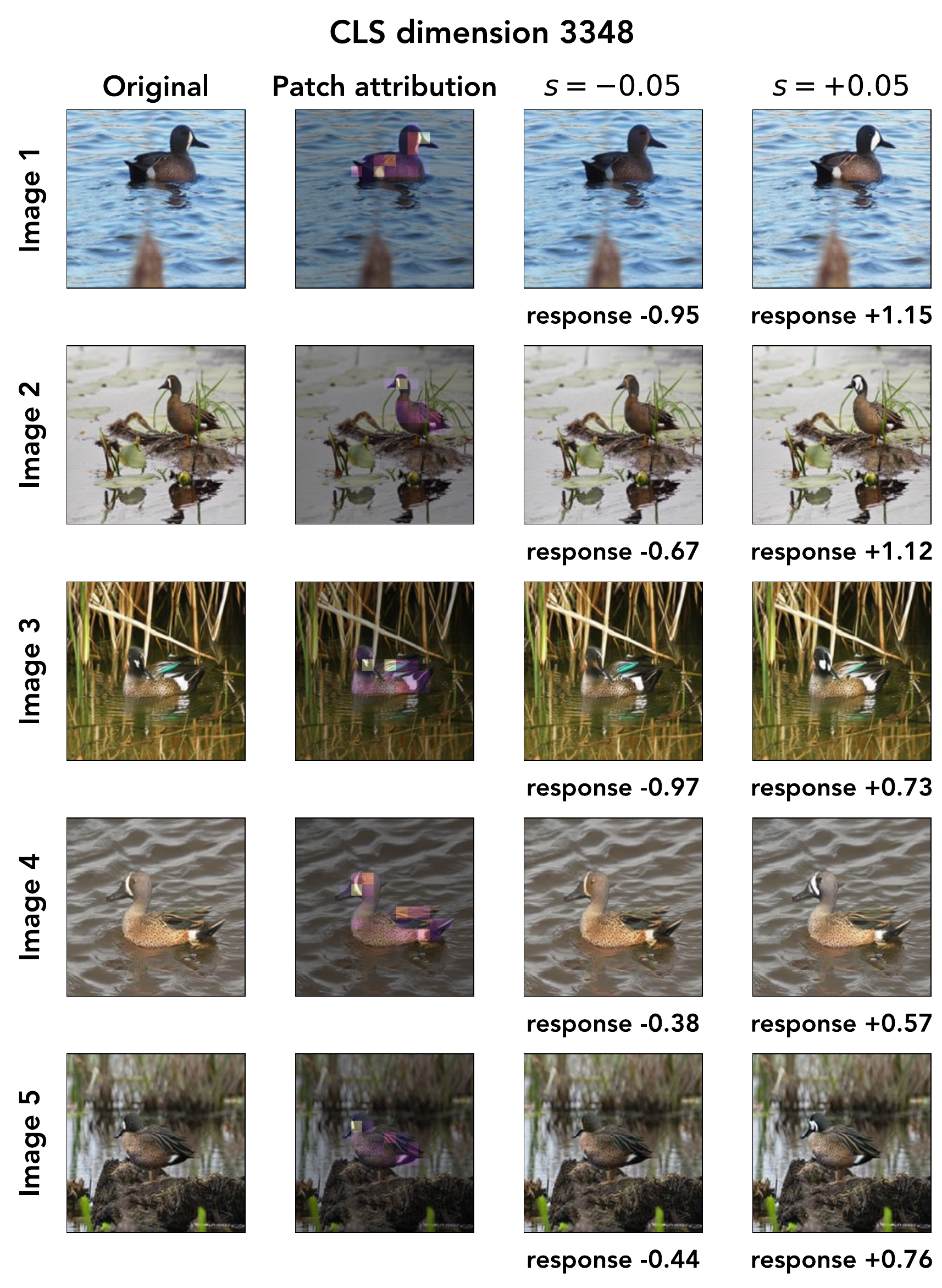}
    \caption{}
\end{subfigure}
\caption{\textbf{Intervention reconstruction consistency results} on \class features (a) $2434$ and (b) $3348$ of NABirds extracted with DINOv3.}\label{fig:five-sample-consistency}
\end{figure}

We present more examples of intervening on \class SAE dimensions for multiple images in \autoref{fig:five-sample-consistency}. The color of the attributed patches becomes lighter as the steering strength increases for dimension $2434$. In comparison, dimension $3348$ focuses on a white strip on the duck's face. Amplifying the feature will make the strip more prominent. These examples prove that the intervention leads to consistent changes in the target \class feature.

\end{document}